\documentclass{article}

\usepackage[final]{neurips_2025}

\usepackage[utf8]{inputenc} % allow utf-8 input
\usepackage[T1]{fontenc}    % use 8-bit T1 fonts
\usepackage{hyperref}       % hyperlinks
\usepackage{url}            % simple URL typesetting
\usepackage{booktabs}       % professional-quality tables
\usepackage{amsfonts}       % blackboard math symbols
\usepackage{nicefrac}       % compact symbols for 1/2, etc.
\usepackage{microtype}      % microtypography
\usepackage{xcolor}         % colors
\usepackage{graphicx}
\usepackage{subcaption} % for subfigures and separate sub-captions

\usepackage{array}
\usepackage{xspace}
\usepackage{tabularx} 
\usepackage{float}
\newcommand{\dataname}{\textsc{NaturalReasoning}\xspace}
\usepackage[most]{tcolorbox}
\usepackage[english]{babel}
\usepackage{enumitem}

\renewcommand\sup[1]{$^{#1}$}

\title{\dataname :  Reasoning in the Wild with 2.8M Challenging Questions}

\author{Weizhe Yuan\sup{1,2}~~~~~~~Jane Yu\sup{1*}~~~~~~~Song Jiang\sup{1}\thanks{\hspace{.15em}Equal contribution.}~~~~~~~
Karthik Padthe\sup{1}~~~~~~~Yang Li\sup{1}\\
\textbf{Ilia Kulikov\sup{1}~~~~~~~Kyunghyun Cho\sup{2}~~~~~~~Dong Wang\sup{1}~~~~~~~Yuandong Tian\sup{1}}\\
\textbf{Jason Weston\sup{1,2}~~~~~~~
Xian Li\sup{1}\thanks{\hspace{.15em}Correspondence to: Xian Li <xianl@meta.com>.}} \\
\\
\textsuperscript{1}Meta~~~~~~~~~~~~~~~
\textsuperscript{2}New York University
}

\begin{document}

\maketitle

\begin{abstract}

Scaling reasoning capabilities beyond traditional domains such as math and coding is hindered by the lack of diverse and high-quality questions. To overcome this limitation, we introduce a scalable approach for generating diverse and challenging reasoning questions, accompanied by reference answers. We present \dataname, a comprehensive dataset comprising 2.8 million questions that span multiple domains, including STEM fields (e.g., Physics, Computer Science), Economics, Social Sciences, and more. We demonstrate the utility of the questions in \dataname through knowledge distillation experiments which show that \dataname can effectively elicit and transfer reasoning capabilities from a strong teacher model. 
Furthermore, we demonstrate that \dataname is also effective for unsupervised self-training using external reward models or self-rewarding. To foster future work, we publicly release \dataname at \url{https://huggingface.co/datasets/facebook/natural_reasoning}.
\end{abstract}

\section{Introduction}
\label{intro}

 Large language models (LLMs) have demonstrated increased reasoning capabilities~\citep{openai2024openaio1card,guo2025deepseek}. % to solve complex tasks. 
 %A notable example is OpenAI's o1 models~\citep{openai2024openaio1card}, which have demonstrated significant improvements in challenging reasoning benchmarks such as AIME\footnote{\url{https://artofproblemsolving.com/online}}, Codeforces~\footnote{\url{https://codeforces.com/}}, and GPQA~\citep{rein2024gpqa}. 
 These models are designed to devote more time to deliberation before responding, enabling them to tackle intricate tasks and solve more complex problems in science, coding, and mathematics. Such reasoning models are trained via large-scale reinforcement learning on tasks where the reward can be derived using rule-based verification. Existing reasoning datasets are often limited to narrow domains where the solutions are short and easy to verify, while the majority of reasoning problems across broader domains are open-ended reasoning. To bridge this gap, we introduce \dataname, a comprehensive dataset curated from pretraining corpora, comprising 2.8 million reasoning questions spanning various topics, including Mathematics, Physics, Computer Science, Economics \& Business, etc. \dataname is compared to a wide range of reasoning datasets, showcasing its 
%beneficial 
advantageous
properties, in particular its
%exceptional 
diversity and difficulty.

\dataname possesses several desirable attributes as a dataset, serving multiple research purposes. Firstly, the questions are backtranslated from the pretraining corpora, ensuring that it represents \textbf{diverse} reasoning problems in the real world, as opposed to synthetic datasets derived from benchmark datasets like MetaMathQA~\citep{yumetamath} and OpenMathInstruct-2~\citep{toshniwal24openmathinstruct}. Secondly, it consists of both questions with \textbf{easy-to-verify} answers and those with \textbf{open-ended} solutions (e.g., theorem proving), providing a rich resource for developing learning algorithms to enhance LLMs' reasoning abilities across broader domains.%beyond a narrow subset of verifiable questions.  
Thirdly, we show that \dataname poses more \textbf{difficult} reasoning problems than existing datasets. Its questions therefore provide an effective testbed for improving LLM reasoning—whether through knowledge distillation from a stronger teacher model or reinforcement learning with external and self-generated reward signals \citep{yuanself}. Lastly, the \dataname dataset complements existing reasoning datasets in terms of both \textbf{quality} and \textbf{quantity}.

Our contributions are threefold:
\begin{itemize}
    \item We create a large-scale and high-quality reasoning dataset by using pretraining data and LLMs alone without extra human annotation. The dataset contains challenging questions which require deliberate thinking accompanied with reference answers. We release the \dataname dataset to the research community at \url{https://huggingface.co/datasets/facebook/natural_reasoning}.
    % to be used as solution verification. 
    \item We show that \dataname is a highly performant dataset to enhance reasoning capabilities in post-training. Specifically, using questions from \dataname in distillation is more sample efficient than existing datasets.% such as OpenMathInstruct-2 and MAmmoTH2~\cite{}.
    \item We investigate how \dataname supports self-training methods. %, including RRFT and DPO, using the model’s own responses. 
    Our results show that the questions in our dataset effectively enable self-learning, where self-rewarding techniques can yield performance comparable to some strong external reward models.
    
    % We also investigate how to curate reliable and verifiable gold answers without human annotations. We compare deriving reliable reward signals using self-consistency and self-scoring, and demonstrate that even without reliable gold answers, the model can still self-improve through self-scoring and curate contrastive pairs for DPO training.
\end{itemize}

\begin{figure*}[!t]
     \centering
    \includegraphics[width=1.0\linewidth]{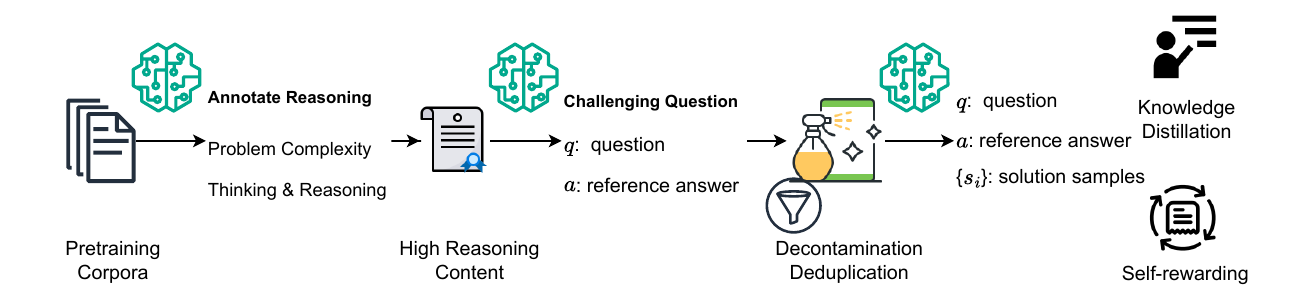}
    \caption{An overview of the data creation approach of \dataname. 
    % First, we use an LLM to annotate the reasoning contents of documents in pretraining corpora. For documents which contains reasoning traces to solve complex problems, we use an LLM to synthesize a self-contained question which requires thinking and reasoning. We also prompt the LLM to return a reference answer to the question if the answer can be found in the document. We then conduct deduplication and decontamination. Sampling solutions to those questions from different LLMs enables knowledge distillation as well as self-rewarding training schemes.
    }
    \label{fig:data_pipeline}
\end{figure*}

\section{Data Collection}

Backtranslating questions based on pretraining corpora has been shown to be a scalable approach to create instruction-following datasets~\citep{li2023self,yue2024mammoth2}. We take a similar approach to extract diverse and realistic reasoning questions grounded in pretraining data, i.e. we generate {\em grounded synthetic data}. An overview of the data creation approach is illustrated in \autoref{fig:data_pipeline}. A key differentiator of our approach is its emphasis on simplicity; we use LLMs throughout the entire cycle of synthetic data generation and curation, without any human annotation nor manual steps such as preprocessing to detect relevant documents and extracting questions with rule-based filtering.

\subsection{Annotate Reasoning}
\label{annotation_sec}
Inspired by the meta-cognitive capabilities of state-of-the-art LLMs~\citep{didolkar2024metacognitive}, we use an LLM to perform detailed annotation of pretraining documents to detect documents which contain sophisticated reasoning traces. We use the public pretraining corpora DCLM-baseline ~\citep{li2024datacomp} and FineMath ~\citep{allal2025smollm2smolgoesbig} as sources, which have demonstrated steeper scaling laws than alternative corpora. More specifically, given a document $d$ from the pretraining corpus, we prompt an LLM to rate the content in $d$ along multiple axes: Problem Completeness, Problem Complexity and Technical Depth, Technical Correctness and Accuracy, Thinking and Reasoning. The detailed prompt is provided in \autoref{app:prompts}. Empirically, we found that the model could analyze the document well and is able to follow the format requirement in the instruction. 
%Please refer to \autoref{app:prompts} for the detailed prompt we use. 

\begin{table*}[t]
\footnotesize
\setlength{\tabcolsep}{5pt}
\renewcommand{\arraystretch}{1.02}

 \caption{Comparison of four large publicly available reasoning datasets with \dataname. ``Q'' denotes ``question'', and question length is measured by the number of words.}
  \label{tab:data_compare}
  
\begin{tabular}{p{2.cm}m{0.9cm}m{0.6cm}p{1cm}p{4cm}p{3.4cm}}
\toprule
\textbf{Dataset}   & \textbf{Domain} & \multicolumn{1}{l}{\textbf{Q. \#}} & \textbf{Q. Len.} & \textbf{Question source}                                & \textbf{Response model}             \\
\midrule
MetaMathQA         & Math            & 395K                                       & 41$\pm$24                 & Rephrased by GPT-3.5-Turbo from GSM8K+MATH              & GPT-3.5-Turbo           \\
NuminaMath         & Math            & 860K                                       & 48$\pm$32                 & Grade-level and competition-level math datasets & GPT-4o     \\
OpenMath2 & Math            & 607K                                       & 46$\pm$44                 & Synthesized by Llama3-405B-Inst from GSM8K+MATH  & Llama3-405B-Inst \\
WebInstruct           & Multi       & 13M                                    & 34$\pm$28                 & Recalled and Extracted from the Web                           & Mixtral-22B×8, Qwen-72B   \\

\midrule
{\small NaturalReasoning}  & Multi       & 2.8M                                     & 55$\pm$21                 & Synthesized by Llama-3.3-70B-Inst grounded in the Web                             & Llama-3.3-70B-Inst \\
\bottomrule
\end{tabular}
\end{table*}

%\subsection{Synthesize Questions, Answers and Responses}
\subsection{Synthesize Questions and Answers}

For documents which are identified with a high degree of reasoning (e.g., scored full points on axes of Problem Complexity and Technical Depth, Thinking and Reasoning), we further prompt an LLM to compose a self-contained and challenging reasoning question $q$ based on the content in $d$. Different from existing work, which extracts questions appearing in the document~\citep{yue2024mammoth2}, our approach allows us to synthesize more novel questions not directly contained in pretraining corpora. Then, we prompt the LLM to verify whether a correct reference answer $a$ to the synthesized question $q$ can be derived from $d$ and, if possible, include it as a reference answer. Finally, for every question we generate an additional response with a relatively strong open-source model (Llama-3-70B-Instruct), which we later use as a teacher signal for knowledge distillation (see \autoref{sec:scaling}).

% In addition to questions and reference answers, we also augment each question in our dataset with a response generated by a relatively strong open-sourced model, Llama3.3-70B-Instruct. These generated responses will be utilized for knowledge distillation, as discussed in \autoref{sec:scaling}.

% the detailed prompts we use are shown in \autoref{app:prompts}.

\subsection{Question Deduplication and Decontamination}
% \wz{Mention that despite question and response, we also have extracted reference answer, and annotated quality/difficulty}
\begin{itemize}[label={}, leftmargin=0pt, labelsep=0pt, itemsep=1pt, labelwidth=0pt]
\item \textbf{Deduplication}
Our deduplication process focuses on identifying and removing near-duplicate questions using locality-sensitive min-hashing at the word level. We apply a similarity threshold of 0.55 to filter out closely related variations, ensuring that questions with the same core reasoning task but different prompts are not redundantly included.

\item \textbf{Decontamination}
We filter out questions that are similar to popular reasoning benchmarks including MATH~\citep{hendrycksmath2021}, GPQA, MMLU-Pro~\citep{wang2024mmlu} and MMLU-Stem~\citep{hendryckstest2021}. The standard 13-gram decontamination method from EleutherAI's llm-eval-harness~\citep{eval-harness} is used to identify and remove $0.026\%$ items from the dataset. 
%Due to their small number, our ablation shows that the effect of these items on SFT evaluation metrics is negligible.
\end{itemize}
% \paragraph{Reasoning Score.}

% \subsection{Bootstraping Reference Answer}
% \wz{I thought we only kept the backtranslated refernece answer as it is...} \xian{These sections are describing the methods, which we use to verify and compare the quality of extracted answer.}
% For each question $q$, we sample $K=32$ solutions $\{a_k\}$ from different models $\{M_i\}$. 

% \subsection{Generate Response}
% \wz{We use Opensourced SOTA LLMs for response generation. Ablation for 405B vs. 3.3-70B vs backtranslation from human CoT}
% \paragraph{Self Consistency.}

% \paragraph{Self Scoring.}

\section{Data Analysis}

We compare \dataname to the following representative existing datasets which were curated to boost reasoning capabilities.

\begin{itemize}[label={}, leftmargin=0pt, labelsep=0pt, itemsep=1pt, labelwidth=0pt]

\item \textbf{MetaMathQA} is created by bootstrapping mathematical questions from GSM8K and MATH, rewriting them from multiple perspectives to form a new dataset~\citep{DBLP:conf/iclr/YuJSYLZKLWL24}. The responses are generated using GPT-3.5-Turbo.

\item \textbf{NuminaMath} is a comprehensive collection of 860K pairs of math problems and solutions~\citep{numina_math_datasets}. The questions cover multiple sources including grade-level questions and competition problems. The solutions in NuminaMath dataset are generated or rewritten by GPT-4o.

\item \textbf{OpenMathInstruct-2} is a collection of 14M synthesized math questions and solutions, based on GSM8K and MATH~\citep{toshniwal2024openmath2}. The solutions are generated by Llama3.1-405B-Instruct~\citep{grattafiori2024llama3herdmodels} and curated through majority vote on the final answer.

\item \textbf{WebInstruct} recalls relevant documents from Common Crawl using a fastText model trained on a diverse seed dataset of quiz websites. It then extracts question-answer pairs contained in recalled web pages and uses LLMs (Qwen-72B~\citep{bai2023qwentechnicalreport}, Mixtral-8×7B~\citep{jiang2024mixtralexperts}) to refine the answer~\citep{yue2024mammoth2}.

\end{itemize}

In addition, we compare models trained on \dataname with those trained on the OpenThoughts dataset~\citep{guha2025openthoughtsdatarecipesreasoning}, a recent open-source collection designed for reasoning in math, code, and science. As shown in \autoref{app:compare_openthoughts}, models trained on \dataname achieve better performance across general reasoning benchmarks, demonstrating its broader coverage and stronger generalization compared to OpenThoughts.

% \paragraph{MetaMathQA} is created by bootstrapping mathematical questions from GSM8K and MATH, rewriting them from multiple perspectives to form a new dataset~\citep{DBLP:conf/iclr/YuJSYLZKLWL24}. The responses are generated using GPT-3.5-Turbo.

% \paragraph{NuminaMath} is a comprehensive collection of 860K pairs of math problems and solutions~\citep{numina_math_datasets}. The questions cover multiple sources including grade-level questions and competition problems. The solutions in NuminaMath dataset are generated or rewritten by GPT-4o.

% \paragraph{OpenMathInstruct-2} is a collection of 14M synthesized math questions and solutions, based on GSM8K and MATH~\citep{toshniwal2024openmath2}. The solutions are generated by Llama3.1-405B-Instruct~\citep{grattafiori2024llama3herdmodels} and curated through majority vote on the final answer.

% \paragraph{WebInstruct} recalls relevant documents from Common Crawl using a fastText model trained on a diverse seed dataset of quiz websites. It then extracts question-answer pairs contained in recalled web pages and uses LLMs (Qwen-72B~\citep{bai2023qwentechnicalreport}, Mixtral-8×7B~\citep{jiang2024mixtralexperts}) to refine the answer~\citep{yue2024mammoth2}.

% \begin{itemize}
%     \item NuminaMath~\citep{numina_math_datasets}.
%     \item OpenMathInstruct-2~\citep{toshniwal2024openmath2}.
% \end{itemize}

\subsection{Basic Statistics}
We present a comparison of key dataset statistics in \autoref{tab:data_compare}. Most large open reasoning datasets primarily focus on the math domain, with datasets such as OpenMathInstruct-2, NuminaMath, and MetaMathQA containing only math-related questions. In contrast, \dataname covers reasoning problems from more diverse domains. Additionally, \dataname consists of 2.8M \textbf{unique} questions, significantly larger than OpenMathInstruct-2 (607K), NuminaMath (860K), and MetaMathQA (395K), though smaller than WebInstruct (13M).

With an average length of 55 words, questions in \dataname are notably longer than those in OpenMathInstruct-2 (46), WebInstruct (34), NuminaMath (48), and MetaMathQA (41). Longer questions embed richer context and multi-step requirements, demanding deeper reasoning. Coupled with our varied, web-grounded sources, this added complexity sets \dataname apart as a uniquely challenging dataset.

% Moreover, our questions have an average length of 55 words, which is longer than those in OpenMathInstruct-2 (46), WebInstruct (34), NuminaMath (48), and MetaMathQA (41). The increased question length suggests that our dataset contains more complex and detailed problems, potentially requiring more advanced reasoning capabilities. This, combined with our diverse question sources and grounding in web-based knowledge, highlights the uniqueness and potential difficulty of our dataset.

% \wz{Num of questions, question length: mean+-std, question similarity (use the embedding to calculate on average, how similar is a question to other questions in the dataset)}
\subsection{Question Quality and Difficulty}

% in the body of your paper

\begin{figure}[t]
    \centering
    \includegraphics[width=0.95\linewidth]{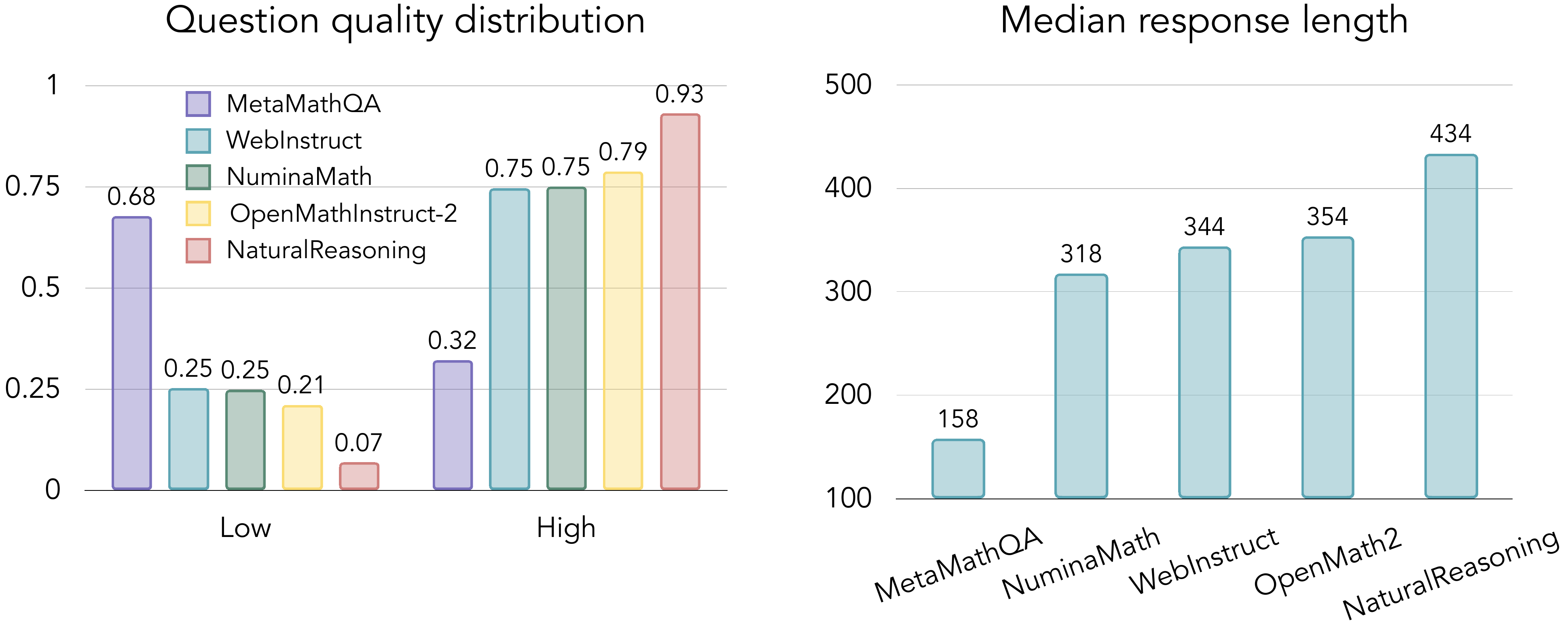}
    \caption{\textbf{Left}: Question quality distribution based on LLM annotations: Low (0-6), High (7-10). \textbf{Right}: Median response length (in words) of Llama3.3-70B-Instruct responses across all datasets. 
    % Responses to questions in \dataname are generally much longer, suggesting more challenging/complex nature.
    }
    \label{fig:difficulty_distribution}
\end{figure}

\paragraph{Quality}
We evaluate question quality in \dataname with both automatic and human judgments. Three strong LLMs (DeepSeek-R1-Distill-Qwen-32B, Qwen2.5-72B-Instruct, Llama-3-70B-Instruct) independently score every question on a 0–10 scale reflecting solvability and completeness. For each comparison dataset, we randomly sample 10\% of its questions and apply the same prompt. Scores of 0–6 are labeled low quality and 7–10 high quality; a question is deemed high quality when at least two models assign a high score. As shown in \autoref{fig:difficulty_distribution}, \dataname contains the highest fraction of high-quality questions (93\%), surpassing the next-best dataset (79\%). To corroborate these results, we conduct a small human study: two expert annotators independently score 100 randomly selected questions from each dataset, and we average their ratings. The pattern holds—\dataname achieves the top mean score (6.45) versus 5.92 for the second best. Full details are provided in \autoref{app:human_eval}.

% The results, presented in \autoref{fig:difficulty_distribution}, show that \dataname has the highest proportion of high-quality questions at 98\%, while the second-highest dataset contains 87\% high-quality questions.

% To estimate question difficulty, we use an off-the-shelf LLM to generate responses and measure their lengths, as longer responses generally indicate more complex questions. Specifically, we employ Llama3.3-70B-Instruct to generate answers for all questions for fair comparison. For the comparison datasets, we analyze responses for a randomly selected 10\% subset got using the same model. The results, shown in \autoref{fig:response_length}, indicate that our dataset has the longest median response length (434 words), significantly exceeding other datasets. The second-longest median response length is observed in OpenMathInstruct-2, which has a median of xx words. This suggests that our dataset contains more intricate and demanding questions compared to existing open reasoning datasets.

\paragraph{Difficulty} To estimate question difficulty, we leverage a strong LLM to generate responses and use response length as a proxy, as longer chain-of-thoughts typically correspond to more complex questions. Specifically, we randomly selected 10\% of questions from each dataset, and employ Llama3.3-70B-Instruct to generate responses for each question. As is shown in \autoref{fig:difficulty_distribution}, \dataname exhibits the longest median response length (434 words), significantly surpassing all other datasets. 
% The second-longest median response length is observed in OpenMathInstruct-2, with a median of 354 words. 
This suggests that our dataset contains more intricate and reasoning-demanding questions compared to existing open reasoning datasets.

\subsection{Question Diversity}
In addition to being difficult, questions in \dataname are also diverse. We analyze diversity of the questions in terms of question similarity and the topics, and compare to WebInstruct, an existing dataset covering multiple domains.
%\wz{1. plot all data's embedding together. 2. show the different topics in our dataset }

\begin{figure*}[t]
  \centering
  % left panel ────────────────────────────────────────────────────────────
  \begin{subfigure}[t]{0.59\textwidth}   % 0.59 leaves a whisker of space
    \centering
    \includegraphics[width=\linewidth]{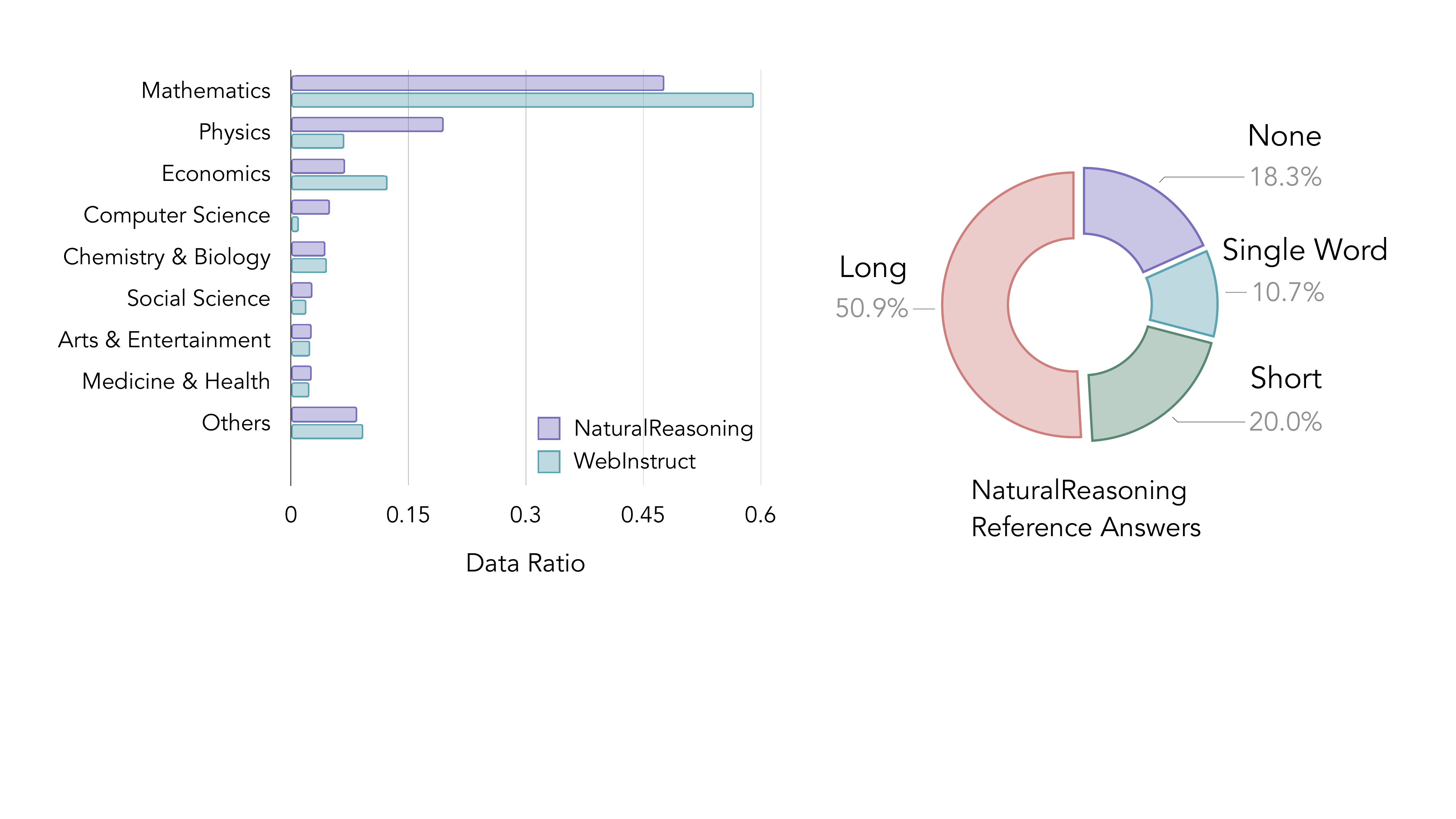}
    \caption{Question topic distributions.}
    \label{fig:topic_dists}
  \end{subfigure}
  \hfill
  % right panel ───────────────────────────────────────────────────────────
  \begin{subfigure}[t]{0.38\textwidth}
    \centering
    \includegraphics[width=\linewidth]{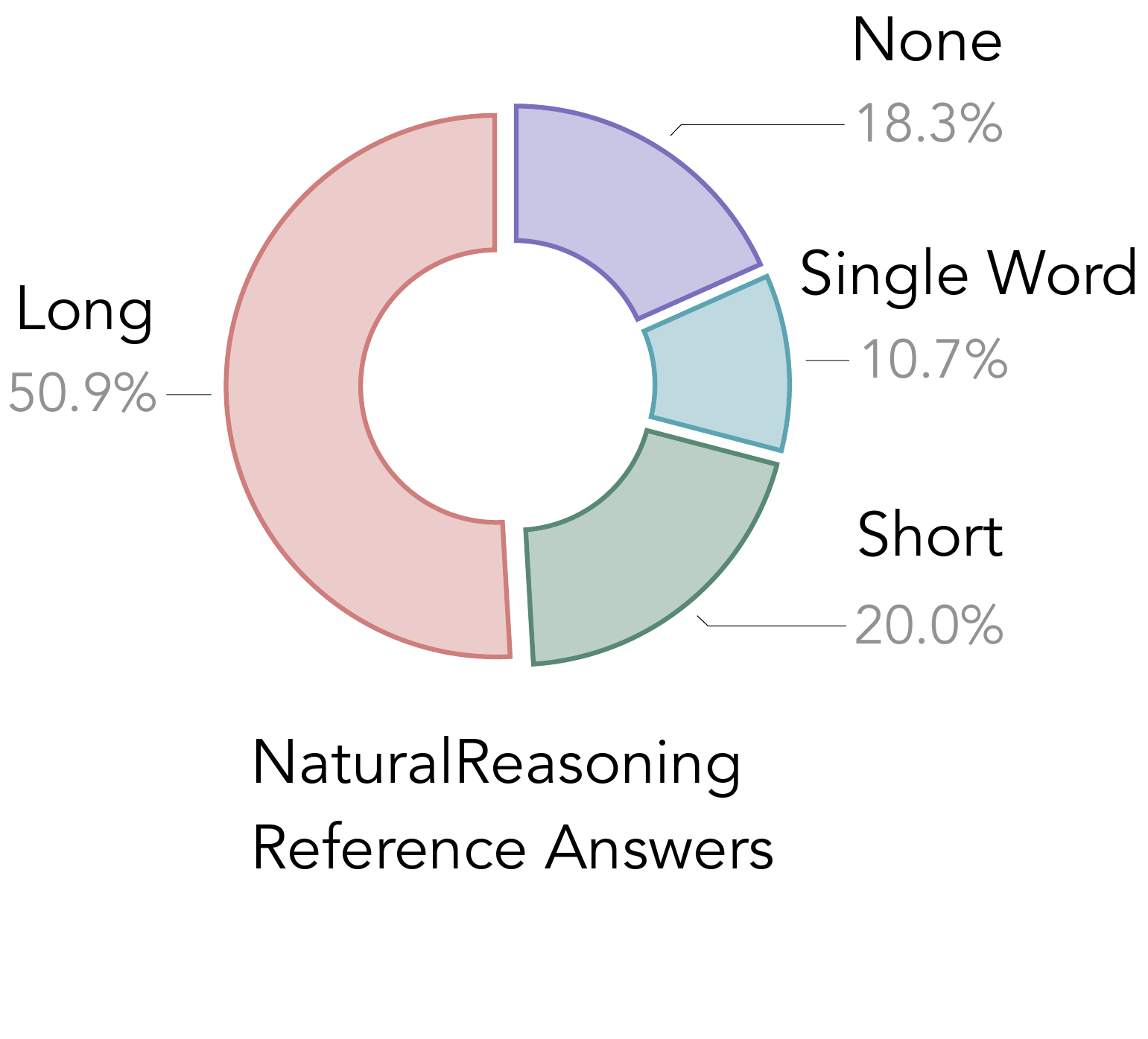}
    \caption{Reference answer lengths in \dataname.}
    \label{fig:answer_lengths}
  \end{subfigure}

  \caption{\textbf{Left}: Topic distributions of \dataname and WebInstruct. \dataname generally shows equivalent or even greater coverage on non-Math topics like Computer Science and Physics. \textbf{Right}: Distribution of reference answer lengths in \dataname, showing that the majority of questions have long reference answers ($\geq$10 words).}
  \label{fig:topic_ref_answer}
\end{figure*}

\paragraph{Embedding Clustering} 
\label{sec:clustering}
% For this method we use LLM to calculate embeddings for sample of MAmmoTH2 and \dataname, then UMAP~\cite{mcinnes2018umap} is used to project high dimensional data to 2D embeddings, then apply K-means to cluster these embedding and then we use 8B scale LLM to label clusters into highlevel categories by passing few examples from the cluster in prompt. From the \ref{fig:topic_clusters} we can see that along with Math data \dataname contains more dense representation from diverse non-math topics like Physics, Chemistry, computer science, law etc. Whereas MAmmoTH mainly skews towards Math content.
We use an off-the-shell sentence encoder\footnote{\url{https://huggingface.co/sentence-transformers/all-MiniLM-L6-v2}} to generate embeddings for questions in WebInstruct and \dataname. We then apply UMAP~\citep{mcinnes2018umap} to project the high-dimensional embeddings into a 2D space, followed by K-means clustering~\citep{wu2012advances} to identify distinct groups. We use Mixtral-8B to assign high-level labels to these clusters, which is prompted with a few examples from each cluster. As \autoref{fig:topic_clusters} shows, \dataname contains a more diverse and dense representation of non-mathematical topics, including Physics, Chemistry, Computer Science, and Law, besides Math. In contrast, WebInstruct is primarily skewed toward mathematical content, highlighting the broader topic coverage of \dataname.
\paragraph{Classifier Categorization} To estimate the topic distribution, a multi-class topic classifier is used to classify each question into 16 knowledge classes. The class labels are motivated by Wikipedia academic disciplines\footnote{\url{https://en.wikipedia.org/wiki/Outline_of_academic_disciplines}}. \autoref{fig:topic_dists} shows that \dataname is complementary to WebInstruct, where \dataname has better coverage on non-Math topics especially Physics, Computer Science, Social Science, etc.

\subsection{Reference Answer Analysis}

Among the 2.8 million questions we synthesized, 81.68\% have reference answers which could be derived from pretraining data. The distribution of reference answer lengths is illustrated in \autoref{fig:answer_lengths}, with single-word answers accounting for 10.7\%, short answers (2–9 words) making up 20.0\%, and long answers ($\geq$10 words) constituting the majority at 50.9\%.

We provide examples of questions with single-word, short, and long answers in \autoref{app:example_questions}. In general, we found that questions with single word answers typically involve numerical, factual, or definitional queries, while questions with long answers demand more free-form in-depth analysis. For questions with a long answer, the extracted reference answer is typically a short summary content from the original documents or useful clues to answer the question. While reference answers may contain some noise, we demonstrate their utility in \autoref{sec:ref_ans_use} for both filtering training data in knowledge distillation and enabling reinforcement learning with verifiable rewards (RLVR)~\citep{lambert2025tulu3pushingfrontiers}.

\section{Experimental Setup}

%\subsection{Training data}
We highlight the efficacy of \dataname in two settings: (1) \textbf{Knowledge distillation}, and (2) \textbf{Unsupervised Self-Training}. 
For (1), we evaluate whether \dataname enables steeper scaling than existing datasets when distilling reasoning capabilities to a student model via supervised finetuning (\autoref{sec:scaling}. We experiment with different model families such as Llama3.1-8B and Qwen2.5-7B. Specifically we show that questions from \dataname are very effective at distilling long chain-of-thoughts from reasoning models and compare it to manually curated questions such as LIMO~\citep{ye2025limoreasoning} and S1K~\citep{muennighoff2025s1simpletesttimescaling} (\autoref{sec:long_cot}).
%of super a student model.  via supervised finetuning, which enables steeper scaling trends than existing reasoning datasets, and (2) using \dataname as a source to extract seed data to target a specific domain at hand. 
To demonstrate (2), we evaluate how well \dataname supports self-training either through a strong external reward model or self-rewarding mechanisms~\citep{yuan2024selfrewarding} (\autoref{sec:unsup_self_train}).

\paragraph{Evaluation} We evaluate our models on a diverse set of benchmarks that encompass both math and science reasoning: MATH, GPQA, GPQA-Diamond~\citep{rein2024gpqa}, and MMLU-Pro. In \autoref{app:cross_domain_generalization}, We also show \dataname's utility for broader NLP tasks (e.g., writing). To ensure a fair and consistent comparison, we adopt a zero-shot evaluation setting across all trained models. For inference we use vllm \citep{Kwon_2023} and employ greedy decoding to maintain determinism and eliminate variability introduced by stochastic generation. Unless mentioned otherwise, we report accuracy averaged over the last three saved model checkpoints during training.

% \paragraph{GPQA.} We evaluate on the Diamond subset.

\section{Steeper Scaling with Challenging and Diverse questions}
\label{sec:scaling}

% in the body
% ─── figure-and-table combo ─────────────────────────────────────────────

% \begin{figure}
%     \centering
%     \includegraphics[width=0.5\linewidth]{fig/scaling_curve_avg.pdf}
%     \caption{Average performance across MATH, GPQA, and MMLU-Pro when using varying sizes of data for training Llama3.1-8B-Base. SFT with 1.5 million examples from \dataname is able to surpass Llama3.1-8B-Instruct.}
%     \label{fig:scaling_base}
% \end{figure}

Our hypothesis is that challenging and diverse questions which require thinking and reasoning are more sample efficient for post-training. To verify this, we run supervised finetuning (SFT) starting from a base model, and evaluate overall performance across MATH, GPQA, and MMLU-Pro.

% \subsection{From the base model}

% \wz{xian: could you add some training details about those models, like for different data size, the lr and bsz, how frequent is a ckpt saved etc.}
% We finetune Llama3.1-8B-Base using fairseq2 training recipes \cite{balioglu2023fairseq2}, with varying sizes of data (number of unique questions), and compare average performance across MATH, GPQA, and MMLU-Pro. 
We fine-tuned the Llama3.1-8B-Base model and Qwen2.5-7B using fairseq2 training recipes \citep{balioglu2023fairseq2}, exploring the impact of varying dataset sizes. Specifically, we trained models on our dataset and the comparison datasets introduced in \autoref{tab:data_compare}, and evaluated their average performance across three benchmarks: MATH, GPQA, and MMLU-Pro. 
% The responses
% are generated by different models as is described in \autoref{tab:data_compare}. 
For all datasets, we train 3 epochs for each data size, using batch size 128, learning rate $5e^{-6}$, with cosine learning rate schedule where final learning rate is 1\% of peak learning rate. 
% We report the average performance of the last 3 checkpoints during the last 6000 steps of training. 

% Note that we didn't do data decontamination for other datasets.

\subsection{Results}

\begin{figure*}[t]
  \centering
  % left: figure ---------------------------------------------------------
  \begin{subfigure}[t]{0.49\textwidth}
    \vspace{0pt} % <─ ensures true top alignment
    \centering
    \includegraphics[width=\linewidth]{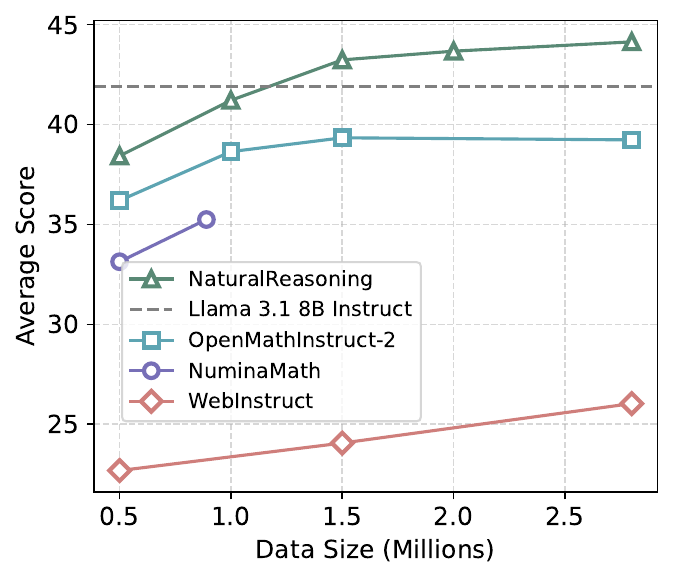}
    \caption{Average performance across MATH, GPQA, and MMLU-Pro when using varying sizes of data for training Llama3.1-8B-Base. SFT with 1.5 million examples from \dataname is able to surpass Llama3.1-8B-Instruct.
    }
    \label{fig:scaling}
  \end{subfigure}
  \hfill
  % right: table ---------------------------------------------------------
  \begin{subtable}[t]{0.49\textwidth}
    \vspace{0pt} % <─ same trick
\footnotesize
\setlength{\tabcolsep}{4pt}
\renewcommand{\arraystretch}{0.95}
\caption{Performance breakdown by benchmark, where the highest accuracy per data size is bolded.
% Math-specific datasets such as NuminaMath and OpenMathInstruct-2 perform well on MATH but poorly on on non-math specific tasks (GPQA and MMLU-Pro). \dataname generalizes well across all three tasks, suggesting high-quality and diverse training data.
}
\vspace{1mm}
\begin{tabular}{lcccc}
\toprule
\textbf{SFT Dataset} & \textbf{500K} & \textbf{1M} & \textbf{1.5M} & \textbf{2.8M} \\
\midrule
\multicolumn{5}{c}{\textit{MATH}} \\
\midrule
WebInstruct              & 9.60  & --     & 11.65 & 14.46 \\
NuminaMath               & 49.53 & --     & --     & --     \\
OpenMathInstruct-2               & \textbf{50.83} & \textbf{54.58} & \textbf{56.47} & \textbf{59.25} \\
NaturalReasoning     & 45.17 & 48.55 & 52.49 & 55.55 \\
\midrule
\multicolumn{5}{c}{\textit{GPQA}} \\
\midrule
WebInstruct              & \textbf{29.02} & --    & 25.37 & 26.12 \\
NuminaMath               & 13.84 & --     & --     & --     \\
OpenMathInstruct-2               & 25.60 & 27.31 & 27.23 & 26.45 \\
NaturalReasoning     & 26.64 & \textbf{29.91} & \textbf{31.77} & \textbf{30.13} \\
\midrule
\multicolumn{5}{c}{\textit{MMLU-Pro}} \\
\midrule
WebInstruct              & 29.44 & --     & 35.17 & 37.54 \\
NuminaMath               & 42.36 & --     & --     & --     \\
OpenMathInstruct-2               & 32.16 & 34.03 & 34.30 & 31.99 \\
NaturalReasoning     & \textbf{43.47} & \textbf{45.16} & \textbf{45.43} & \textbf{46.71} \\
\bottomrule
    \end{tabular}
    
    \label{tab:scaling_breakdown}
  \end{subtable}

  \caption{Scaling results for Llama3.1-8B-Base model.}
  \label{fig:scaling_fig_tab}
\end{figure*}

% As is shown in \autoref{fig:scaling_base}, \dataname is more sample efficient than existing datasets commonly used in reasoning post-training at different scales. More importantly, only training on 1.5 million questions from \dataname already outperforms Llama 3.1 8B Instruct. On the other hand, training using OpenMathInstruct-2 and MAmmoTH2 do not manage to surpass the Llama 3.1 8B Instruct, even with 2.8 million datapoints. A breakdown of these scores is enumerated in Table~\ref{tab:scaling_base_breakdown}, which shows that math-specific datasets like NuminaMath and OpenMathInstruct-2 perform well if evaluated on math-centric datasets like MATH but are much less performant on more general benchmarks like GPQA and MMLU-Pro than \dataname.
The scaling trends for Llama3.1-8B-Base model plotted by averaging performance on the three benchmarks are shown in \autoref{fig:scaling}, and \autoref{tab:scaling_breakdown} provides a detailed breakdown of model performance across different dataset sizes and benchmarks. The scaling trends for Qwen2.5-7B model also show the superiority of \dataname and we put the results in \autoref{app:scale_qwen}.

% \wz{Should we mention that we did not do decontamination for other datasets and numina has MATH and MAmmoTH has MMLU-Pro}\xian{sure.}

% Overall, we make the following observations.

\paragraph{\dataname is significantly more sample-efficient than existing reasoning datasets.} As shown in \autoref{fig:scaling}, models trained on \dataname require fewer training examples to achieve superior performance. With just 1.5 million training examples, the model trained on \dataname already outperforms Llama3.1-8B-Instruct, which was extensively tuned for instruction-following with more data~\citep{grattafiori2024llama3herdmodels}. In contrast, other datasets, including OpenMathInstruct-2 and WebInstruct, fail to surpass Llama3.1-8B-Instruct even when trained on 2.8 million data. Each \dataname sample therefore provides denser, more effective reasoning supervision, making it the most data-efficient choice for boosting model reasoning performance.

\paragraph{Math-specific datasets like OpenMathInstruct-2 excel at math reasoning but fail to generalize beyond math.} A closer look at \autoref{tab:scaling_breakdown} reveals that OpenMathInstruct-2 consistently achieves the highest scores on the MATH benchmark, with performance increasing from 50.83 (500K) to 59.25 (2.8M). This confirms that OpenMathInstruct-2 is well-optimized for pure math reasoning. However, its performance on GPQA and MMLU-Pro is significantly weaker, where GPQA accuracy plateaus around 27–26 as dataset size increases, and MMLU-Pro accuracy fluctuates without significant improvement. This suggests that while OpenMathInstruct-2 provides strong supervision in math reasoning, it lacks the diversity required to generalize to broader scientific reasoning tasks.

\paragraph{Some datasets show diminishing returns as training data increases, highlighting potential inefficiencies in data composition.} While scaling up dataset size generally improves performance, datasets like WebInstruct and OpenMathInstruct-2 exhibit inconsistent or plateauing performance trends. For example, WebInstruct’s GPQA performance peaks at 500K (29.02) but drops at 1.5M (25.37) and only marginally improves at 2.8M (26.12). Similarly, OpenMathInstruct-2’s GPQA accuracy fluctuates with increased training data, suggesting that simply adding more data does not always lead to better reasoning abilities. These observations imply that data quality and diversity matter more than data volume when training models for complex reasoning.

\section{Eliciting Long Chain-of-Thought}
\label{sec:long_cot}

% \begin{figure}[t]
%     \centering
%     \includegraphics[width=0.5\linewidth]{fig/r1_distill.pdf}
%     \caption{
%     % R1-distilled results on GPQA-Diamond. The R1-distilled baseline is obtained by distilling DeepSeek-R1 to finetune Llama-3.3-70B-Instruct. The reported baseline uses a curated set of 800k questions, as reported in \citet{guo2025deepseek} whereas our baseline uses questions from \texttt{OmniReason-15k}. Despite using 50x less data, our R1-distilled model is able to outperform the r baseline.
%     R1-distilled model performance on GPQA-Diamond. Our baseline (distilled from \texttt{OmniReason-15k}) outperforms the reported baseline \cite{guo2025deepseek} (distilled from 800k questions) despite using 50x less data. Results are shown for the reported baseline and our baseline at different maximum token lengths during inference time.
%     }
%     \label{fig:r1_distill}
% \end{figure}

% WEIZHE'S ORIGINAL FIRST PARAGRAPH:
% Following the emergence of OpenAI-O1, Recent studies show that by allowing the model to think for longer by generating extended CoT, it is able to solve questions that it previously cannot be solved using short CoT. To show that questions from \dataname are difficult enough and has the potential to let model think for longer, we try distilling the recent Deepseek-R1 model to Llama3.3-70B-Instruct using SFT. 

In addition to the emergence of OpenAI-o1 and DeepSeek-R1, several studies suggest that simpler tasks require fewer steps while complex tasks benefit  significantly from longer CoTs \citep{jin2024impact}. When encouraging the model to think for longer, they are able to solve questions that they previously could not. Motivated by this, we investigate whether the questions in \dataname are complex enough to benefit from longer CoTs from a stronger reasoning model. We do so by distilling Deepseek-R1 responses to Llama3.3-70B-Instruct.
% under the assumption that performance can greatly benefit from this stronger teacher model. 

% We use our dataset to SFT, obtaining our R1-distilled Llama and compare this to the R1-distilled LLama reported in \cite{guo2025deepseek}, which uses a curated set of 800k reasoning and non-reasoning data.

% 1) our questions benefit more from a stronger reasoning model, 2) distill on our questions are better than 800k questions. 

% We also show that questions from \dataname could elicit long CoT responses which could benefit distilling the reasoning performance from a strong reasoner to a weaker student model. 

% \wz{sry, i'm thinking how could we better motivate this experiment... cite o1 etc} \jane{I tried to write something but now I don't feel like it makes sense...}

We randomly sampled 1K questions from \dataname and used SGLang \citep{zheng2023sglang} to prompt DeepSeek-R1, generating one response per question. Resulting response lengths range from 745 to 14.6K tokens with an average length of 4430 tokens. We then supervised finetune Llama-3.3-70B-Instruct on this set and compare its performance against training on two strong, heavily curated datasets: s1K-1.1 \citep{muennighoff2025s1simpletesttimescaling} and LIMO \citep{ye2025limoreasoning}.
Both datasets underwent multiple filtering stages to ensure that their questions are high-quality, diverse, and challenging; for consistency, all responses were generated with DeepSeek-R1. To keep the evaluation consistent with the setting used in \citet{guo2025deepseek}, we report pass@1 averaged across $n=24$ samples. Each sample is generated using temperature=$0.6$, top\_p=$0.95$. 

The results in \autoref{tab:r1_distill} show that a randomly selected subset of 1k questions from \dataname matches—even slightly exceeds—the performance obtained on datasets that underwent several rounds of meticulous filtering and curation. This parity underscores that \dataname contains questions that are diverse, challenging, and of consistently high quality.

To examine the impact of scale, we expanded the random sample size from 1K to 10K and finally to 100K \dataname questions; the corresponding results are also presented in \autoref{tab:r1_distill}. Performance increases monotonically across every benchmark, providing further evidence that enlarging the slice of \dataname delivers substantial gains precisely because the added questions maintain the same high standard of quality. Moreover, fine-tuning Llama-3.3-70B-Instruct on just 100K randomly sampled questions from \dataname brings the model close to DeepSeek-R1-Distill-Llama-70B, which was trained on 800K examples. Despite using only one-eighth the data, our model outperforms DeepSeek-R1-Distill-Llama-70B on GPQA-Diamond and MMLU-Pro, and falls only slightly behind on MATH-500. This further attests to both the scalability and the intrinsic quality of \dataname.

\begin{table}[t]
\footnotesize
\setlength{\tabcolsep}{3pt}
\renewcommand{\arraystretch}{1}
\caption{
Pass@1 of Llama-3.3-70B-Instruct after distilling DeepSeek-R1 responses. We compare the performance of random selection from \dataname with curated datasets such as s1K-1.1\citep{muennighoff2025s1simpletesttimescaling} and LIMO\citep{ye2025limoreasoning}, as well as the scaling effect of \dataname.
}
\label{tab:r1_distill}
\begin{tabular}{lccccc}
\toprule
             & \textbf{Training size}  & \textbf{GPQA-Diamond} & \textbf{MMLU-Pro} & \textbf{MATH-500} & \textbf{Average} \\
             \midrule
Llama3.3-70B-Instruct & 0 & 50.5 & 70.5 & 77.0 & 66.0 \\
LIMO    & 817 & 56.5         & 76.8      & 86.6  & 73.3 \\
s1K-1.1           & 1,000 & 62.7         & 77.4      & 86.6   & 75.6 \\

\midrule
\dataname & 1,000 & 63.5         & 78.0      & 86.2   & 75.9\\
\dataname & 10,000 & 65.6 & 78.4 & 87.4 & 77.1 \\
\dataname & 100,000 & \textbf{67.3} & \textbf{79.5} & 89.8 & 78.9\\
\midrule
DeepSeek-R1-Distill-Llama-70B & 800,000 & 65.2 & 78.5 & \textbf{94.5} & 79.4 \\

\bottomrule
\end{tabular}
\end{table}

\section{Unsupervised Self-Training}
\label{sec:unsup_self_train}

% \wz{Add RMs, self-scoring, self-consistency to construct DPO pairs}
% \wz{xian could you add some details in the Appendix regarding how we sample the responses?}
Since open-ended reasoning questions are difficult to evaluate using exact match with reference answers, we explore whether our dataset can facilitate self-training through either strong external reward models or self-reward mechanisms~\citep{yuanself}

% self-training strategies based on self-consistency~\cite{prasad2024selfconsistencypreferenceoptimization} and self-reward mechanisms~\cite{yuanself}.  
%we focus on science reasoning benchmarks, such as GPQA. To achieve this, we first sample 250 questions from the GPQA dataset, excluding those from the Diamond subset. For each selected question, we retrieve the 1,000 most similar questions from \dataname, which were already decontaminated against the entire GPQA datset. Similarity is computed using cosine similarity between two question embeddings. After obtaining this candidate set, we apply deduplication and perform clustering, grouping the questions into 15,000 clusters. From each cluster, we select the questions closest to the cluster center, ensuring a diverse and representative dataset for downstream science reasoning tasks. This process resulted in a pool of 15,000 questions, which we refer to as \texttt{OmniReason-15k}. Models trained on this subset can still be evaluated on the GPQA-Diamond test set as those questions are not used for data selection. \texttt{OmniReason-15k} is used in \autoref{sec:ref_ans_use} and \autoref{sec:unsup_self_train}.
To test the effectiveness of self-training without confounding factors from distribution shift, we evaluate on  GPQA-Diamond as test set, and use the remaining questions from GPQA as seeds to retrieve similar questions from \dataname. We curated 15,000 questions in total, which we refer to as \texttt{SelfTrain-15k}. More details are in \autoref{app:self_train_15k}. Models trained on this subset can still be evaluated on the GPQA-Diamond test set as those questions are not used for data selection.

%We use the \texttt{OmniReason-15k} questions and 
%We sample 32 responses from Llama3.1-8B-Instruct for each question in this training set. 
We verify unsupervised self-training under two different training method: Rejection-based sampling Fine-Tuning (RFT) and Direct Preference Optimization (DPO)~\citep{rafailov2023direct}, focusing on the effectiveness of different reward scoring strategies. Each approach relies on sampling 32 candidate responses per question, followed by selecting responses based on reward scores. RFT employs rejection sampling, selecting the highest-scoring response for SFT training, while DPO constructs training pairs using both the highest and lowest-scoring responses. For external reward models, we consider \texttt{Qwen2.5-Math-RM-72B}~\citep{yang2024qwen25mathtechnicalreportmathematical} and \texttt{INF-ORM-Llama3.1-70B}\footnote{https://huggingface.co/infly/INF-ORM-Llama3.1-70B}. In addition, we explore a self-rewarding framework where the model evaluates and assigns rewards to its own generated responses. Specifically, we consider the following self-rewarding strategies:
\begin{itemize}[label={}, leftmargin=0pt, labelsep=0pt, itemsep=1pt, labelwidth=0pt]
\item \textbf{Self-consistency}: Inspired by prior work such as \citet{prasad2024self}, the best response is selected based on response frequency, while the worst response is chosen randomly. To determine frequencies, we extract final answers formatted as \verb|\boxed{|X\verb|}| and compute their occurrence counts. Responses without a clearly extractable final answer are filtered out.
\item \textbf{Self-scoring}: The model receives the question and candidate response in a single prompt and is asked to assess whether the response is valid. We define the reward as the log-probability difference between the judgements ``yes'' and ``no''. The full prompt is in \autoref{tab:prompt_self_score}.
\item \textbf{Self-scoring with filtering}: on top of self-scoring, when applying RFT or DPO, we introduce an additional filtering mechanism. Specifically, for RFT, if the highest-ranked response has a self-score below zero, it is discarded. For DPO, if the preferred response in a pair has a self-score below zero, the pair is removed from training.
\end{itemize}

We train Llama3.1-8B-Instruct using RFT data and DPO data constructed through these methods. We use learning rate of $1e^{-6}$, batch size of 64, and train for three epochs, with checkpoints saved every 50 steps.  We report test performance on GPQA-Diamond and MMLU-Pro in \autoref{tab:unsup_self_train}. 

% We evaluate our models on GPQA-Diamond and MMLU-Pro, taking the average performance across the last three checkpoints as the final reported result.  

\subsection{Results}
\begin{table}[t]
\footnotesize
\setlength{\tabcolsep}{20pt}
\renewcommand{\arraystretch}{1}
\caption{Unsupervised self-training results. We employ RFT and DPO training of Llama3.1-8B-Instruct, using various reward scoring strategies. 
% Self-reward methods generally exceed strategies requiring an external reward model, and are further improved when filtering out samples that are predicted as low-scoring. All self-training strategies improve over the instruction-tuned baseline, showing the utility of our dataset for self-training. 
}
\label{tab:unsup_self_train}
\begin{tabular}{lccc}
\toprule
\textbf{Model}                & \textbf{GPQA-Diamond}                     & \textbf{MMLU-Pro}        & \textbf{Average}                    \\
\midrule
Llama3.1-8B-Instruct & \multicolumn{1}{c}{31.82} & \multicolumn{1}{c}{49.79} & \multicolumn{1}{c}{40.81} \\
\midrule
\multicolumn{3}{l}{\textit{RFT training using external reward model \& self-reward}}                               \\
\midrule
INF-ORM-Llama3.1-70B              &        \multicolumn{1}{c}{32.66}	                          &            \multicolumn{1}{c}{50.95}   & \multicolumn{1}{c}{41.81}                   \\
Qwen2.5-Math-RM-72B                 &   \multicolumn{1}{c}{34.18}                               &   \multicolumn{1}{c}{49.84}           &   \multicolumn{1}{c}{42.01}                     
\\
% \midrule

% \multicolumn{3}{l}{\textit{RFT training using self-reward}}                                         \\
% \midrule
Self-consistency     & \multicolumn{1}{c}{34.18}        & \multicolumn{1}{c}{49.83}    & \multicolumn{1}{c}{41.91}      \\
Self-score           & \multicolumn{1}{c}{34.34} & \multicolumn{1}{c}{50.36}
 & \multicolumn{1}{c}{42.35}   \\
Self-score-filtered           & \multicolumn{1}{c}{\textbf{35.02}} & \multicolumn{1}{c}{50.06}  & \multicolumn{1}{c}{42.54}  
\\
\midrule
\multicolumn{3}{l}{\textit{DPO training using external reward model \& self-reward}}                               \\
\midrule
INF-ORM-Llama3.1-70B              &        \multicolumn{1}{c}{33.50}	                          &            \multicolumn{1}{c}{\textbf{52.74}}       & \multicolumn{1}{c}{43.12}                 \\
Qwen2.5-Math-RM-72B                 &   \multicolumn{1}{c}{30.13}                               &   \multicolumn{1}{c}{49.17}     & \multicolumn{1}{c}{39.65}                          \\

% \midrule
% \multicolumn{3}{l}{\textit{DPO training using self-reward}}                                         \\
% \midrule
Self-consistency     & \multicolumn{1}{c}{30.81}        & \multicolumn{1}{c}{48.60}     & \multicolumn{1}{c}{39.71}     \\
Self-score           & \multicolumn{1}{c}{34.34} & \multicolumn{1}{c}{52.11}  & \multicolumn{1}{c}{43.22}  
\\
Self-score-filtered           & \multicolumn{1}{c}{\textbf{35.02}} & \multicolumn{1}{c}{52.31}  & \multicolumn{1}{c}{\textbf{43.67}}  
\\
\bottomrule
\end{tabular}
\end{table}

% Overall we make the following observations.

\paragraph{Self-training improves performance over the baseline.} Llama3.1-8B-Instruct, serving as the baseline, achieves an average score of 40.81 across GPQA-Diamond and MMLU-Pro. Almost all self-training methods lead to improvements, demonstrating the effectiveness of fine-tuning on high-quality model-generated responses.

\paragraph{Self-reward methods are highly competitive, often surpassing external reward models.} While using external reward models, such as INF-ORM-Llama3.1-70B, could outperform the baseline, self-reward methods achieve comparable or even superior results. Notably, self-score-filtered SFT and self-score-filtered DPO deliver the best performance on GPQA-Diamond (35.02), with self-score-filtered DPO achieving the highest overall score (43.67). These results highlight that self-reward mechanisms can effectively guide self-training without relying on external reward models.

\paragraph{Self-score filtering further enhances performance by improving training data quality.} Among self-reward methods, applying simple filtering improves results across both RFT and DPO. In RFT, self-score-filtered (42.54 AVG) outperforms unfiltered self-scoring (42.35 AVG), while in DPO, self-score-filtered (43.67 AVG) surpasses unfiltered self-scoring (43.22 AVG). This suggests that filtering out low-confidence responses strengthens self-training by reducing noise in the training data.

% DPO using INF-ORM-Llama3.1-70B achieves the highest MMLU$_\text{pro}$ score, but self-reward performs better overall.

% While DPO with INF-ORM-Llama3.1-70B achieves the highest MMLU$_\text{pro}$ score (52.74), it falls short on GPQA$_\text{diamond}$ (33.50). In contrast, DPO with self-score filtering achieves a more balanced performance, obtaining the best overall average (43.67). This suggests that while external reward models can excel in specific benchmarks, self-reward methods provide more consistent improvements across tasks.

% Weizhe did:
% self-scoring with reference COT+judgement
% self-scoring without reference

% logp(yes), logp(no)
% logp(##judgement: yes)
% logp(##judgement: no)

% logp(\boxed{yes})
% logp(yes) -14
% logp(##judgement: yes): -2

% \subsection{Self-Consistency}

% \subsection{Self-scoring}

\section{Related Work}

\paragraph{Synthetic Reasoning Data.} Synthetic data has emerged as a promising solution for improving performance. Some approaches bootstrap new data from existing data (e.g., STaR \citep{zelikman2022star} augments with new CoT rationales and MetaMath \citep{yumetamath} rewrites the questions in MATH and GSM8K in several ways), but these techniques rely on the existence of a high-quality dataset. Other techniques such as that of OpenMathInstruct-2 \citep{toshniwal24openmathinstruct}, Xwin-Math \citep{li2024common}, and Self-Instruct \citep{wang2023self} generate new data from only a few seed examples using an LLM but scaling to new domains remains a significant challenge. MMIQC \citep{liu2024augmenting} parses QA pairs from Mathematics Stack Exchange, using the highest-ranked answer, but few measures are taken to curate for quality and the resulting dataset is also specific to the math domain. Similar to our work, WebInstruct \citep{yue2024mammoth2} harvests question-answer pairs from pre-training corpora and spans multiple domains, but is dependent on carefully crafted rule-based filters.

\paragraph{Unsupervised Self-training} Most prior works typically depend on human-annotated (gold) final answers \citep{zelikman2022star, chen2024self, pang2024iterative} or the use of an external reward model \citep{singh2023beyond, dong2023raft}. However, manually annotating or verifying final answers is particularly resource-intensive for complex, multi-step problems and training effective reward models for reasoning often requires human evaluation of LLM outputs \citep{cobbe2021training, uesato2022solving, lightman2023let}, making it similarly costly. Like works such as \citet{she2024mapo, yuan2024selfrewarding, rosset2024direct, viethoangtranduong}, our work explores self-training in the absence of gold labels and does not limit itself to questions with short, easily verifiable targets.

% Metamath bootstraps (augments)
% GAIR-Abel (reformats)
% Xwin-Math (generate new)
% MMIQC (uses Mathematics Stack Exchange, take highest voted answer)

% Simulated environment for code (Shypula et al 2023, InterCode, )
% Self-instruct (needs seed)
% Code Evol-Instruct 
% OSS-Instruct
% star

\section{Conclusion}
We present \dataname, a dataset of 2.8 million questions for enhancing LLM reasoning capabilities. Our questions are challenging,  requiring more deliberate thinking than existing datasets. The dataset covers diverse reasoning problems across multiple domains including math, physics, computer science, economics, social sciences, etc. Using questions from \dataname in distillation experiments, we observe consistent improvement on reasoning benchmarks when scaling  the data size. 
% Using only 15k examples, distillation with an even more powerful teacher model, DeepSeek-R1, enables superior performance
% We also compare different approaches to evaluate the correctness of solutions.    
% Acknowledgements should only appear in the accepted version.
% \section*{Acknowledgements}
% \textbf{Do not} include acknowledgements in the initial version of the paper submitted for blind review.
We also demonstrate that \dataname is effective for enabling LLM unsupervised self-training using external reward models or self-rewarding.
% We also demonstrate that our reference answers can be useful for further curation and that \dataname is effective for enabling LLM unsupervised self-training even in the absence of reference answers.
\section*{Limitation \& Impact Statement}
Although our study already validates \dataname’s value for large-scale offline training—covering supervised distillation and preference-based self-training (RFT, DPO)—we also conduct preliminary experiments using online RL with verifiable rewards with General Verifier, which show promising gains even with limited training. A more systematic exploration of reinforcement learning paradigms, including alternative reward models and scaling strategies remains natural extensions for future work. This paper seeks to improve reasoning capabilities of large language models through leveraging pretraining corpora. While our efforts are focused on curating high-quality, diverse data, models trained using this data may exhibit undesirable behavior not examined in our work. Therefore, comprehensive evaluation would be needed to evaluate and address any potential pre-existing or existing biases in LLMs which leverage this data.

% \section*{References}

\bibliography{citations}
\bibliographystyle{plainnat}

\clearpage

%%%%%%%%%%%%%%%%%%%%%%%%%%%%%%%%%%%%%%%%%%%%%%%%%%%%%%%%%%%%

\appendix

\section{Clustering Results}

\begin{figure*}[htbp]
  \centering
  % ---- left sub-figure ----
  \begin{subfigure}[t]{0.49\textwidth}
    \centering
    \includegraphics[width=\linewidth]{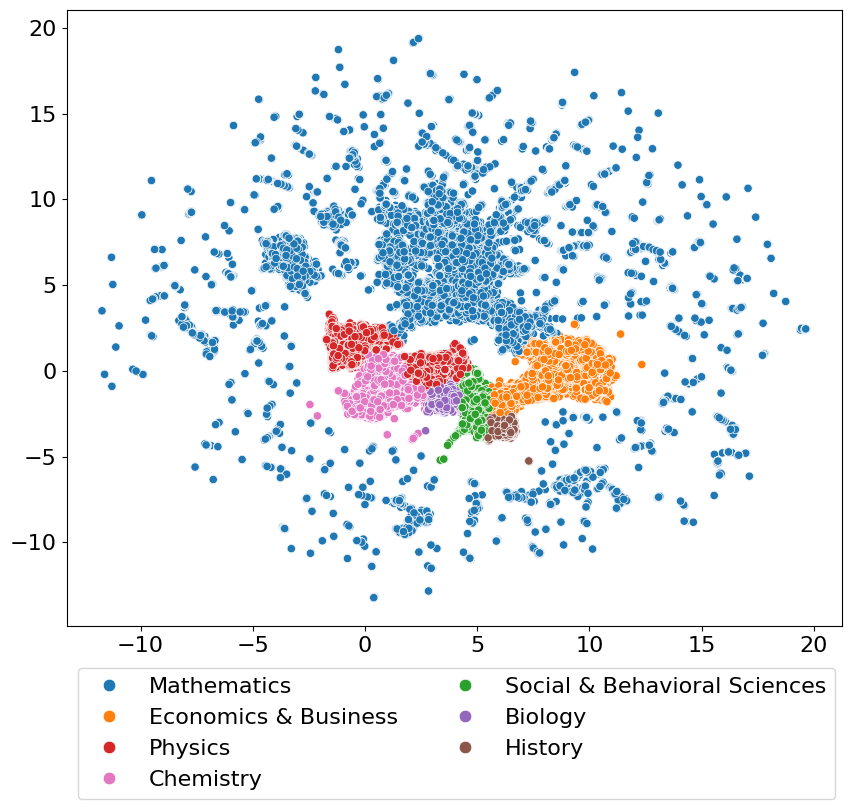}
    \caption{WebInstruct}
    \label{fig:mammoth_topics}
  \end{subfigure}
  \hfill
  % ---- right sub-figure ----
  \begin{subfigure}[t]{0.49\textwidth}
    \centering
    \includegraphics[width=\linewidth]{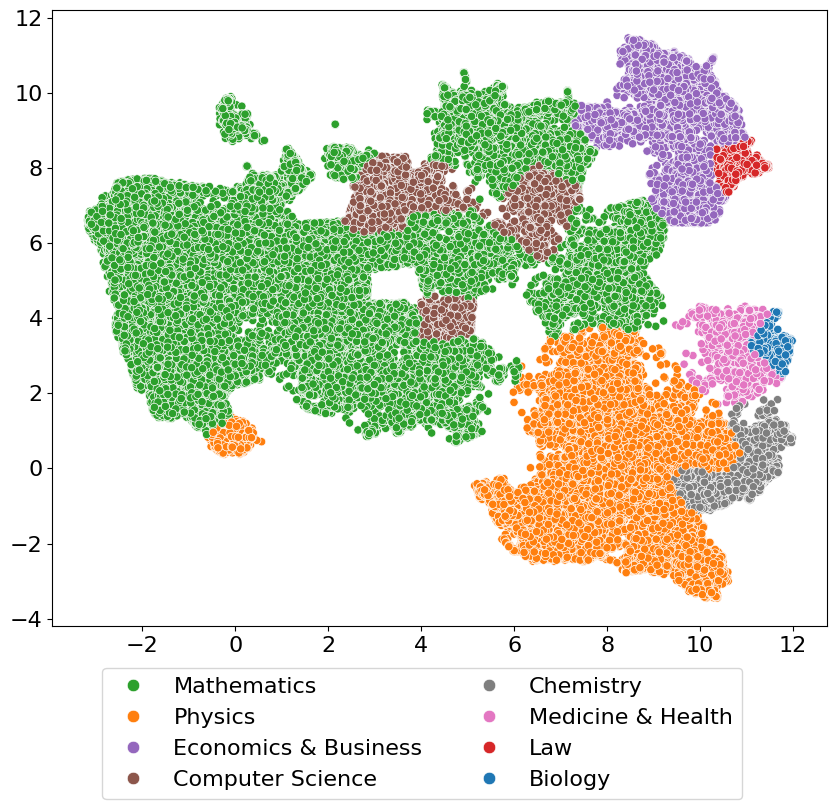}
    \caption{\dataname}
    \label{fig:ourdata_topics}
  \end{subfigure}

  \caption{Topic clustering of WebInstruct and \dataname.}
  \label{fig:topic_clusters}
\end{figure*}

We present the results of our clustering in Figure~\ref{fig:topic_clusters}. The procedure for producing this clustering is described in Section~\ref{sec:clustering}.

% \begin{figure*}[ht]
%     \centering
%     \subfigure[WebInstruct]{\includegraphics[width=0.48\textwidth]{fig/mammoth_topics.png}}
%     % \subfigure[Topic distribution comparison]{\includegraphics[width=0.55\textwidth]{mamoth_ourdata_topic_comparision.png}}
%     \subfigure[\dataname]{\includegraphics[width=0.48\textwidth]{fig/ourdata_topics.png}}
%     % \begin{subfigure}{0.5\textwidth}
%     %     \centering
%     %     \includegraphics[width=0.4\linewidth]{mamoth_topics_kmeans_new.png}
%     %     \caption{MAmmoTH2}
%     %     \label{fig:ourdatacluster}        
%     % \end{subfigure}
%     % \begin{subfigure}{0.5\textwidth}
%     %     \centering
%     %     \includegraphics[width=0.4\linewidth]{ourdata_topics_kmeans_new.png}
%     %     \caption{\dataname}
%     %     \label{fig:openmath_cluster}        
%     % \end{subfigure}
%     \caption{Topic clustering of WebInstruct and \dataname.}
%     \label{fig:topic_clusters}
% \end{figure*}

\section{Example questions}
\label{app:example_questions}
Example questions with single word answer, short answer, long answer are shown in \autoref{tab:example_questions}.

\begin{table*}[!ht]
    \centering
    \footnotesize
  \caption{Example questions with single word, short, and long reference answers.}
  \begin{tabularx}{\textwidth}{@{}X@{}}
    \toprule
    \textbf{Example questions with single word answer} \\ \midrule
    (1) You have 5 cards with letters and numbers on either side. The rule is: If the letter on one side of a card is a vowel, the number on the other side is odd. What is the minimum number of cards you need to turn over to prove or disprove this rule?\\
    (2) What is the approximate number of grapes required to produce one bottle of wine, considering the conversion factors and variability in grape yields, and how does this relate to the overall wine production process? \\
    (3) A company is facing financial difficulties and is in danger of not meeting its obligations. In an effort to secure a bridge loan, the company's management decides to manipulate its financial statements by not reversing cancelled orders, thereby overstating its accounts receivable. This action allows the company to collateralize the loan and secure the necessary funding. However, this practice is in direct violation of revenue recognition rules. Analyze this situation and determine whether the company's actions constitute blatant fraud. Be sure to discuss the technical and ethical implications of such actions. \\
    (4) Evaluate the definite integral $\int_{0}^{\pi} \cos^2(x)\sin^7(x)dx$ using an appropriate substitution method and provide a step-by-step solution.\\
    (5) Speed of boat in still water is 10kmph. If it travels 24km downstream, 16km upstream in the same amount of time, what is the speed of the stream?
     \\
     \midrule
    \textbf{Example questions with short answer} \\ \midrule
    (1) What are the parameters that need to be estimated in the general equation of an ellipse, and how do the variables $x$ and $y$ differ from the constants $a$ and $b$ in this context? Provide a detailed explanation of your answer, including the roles of $a$, $b$, $x$, and $y$ in defining the ellipse.\\
    (2) Given a company's historical data on revenues, working capital, and net capital expenditures, is it acceptable to forecast change in working capital / net capex by regressing (linear) historical data on revenues? What are the limitations of this approach, and what alternative methods could be used to improve the accuracy of the forecast?\\
    (3) Solve the inequality 4(3w+4) $\geq$ 4(2w+12) using interval notation, and express the solution in set-builder notation.\\
    (4) Given an image with shape [1,28,28], what will be the shape of the output of a convolution layer with 10 5x5 kernels (filters) without padding? Assume the image dimensions follow the CHW (Channel, Height, Width) format.\\
     (5) A gas bubble, from an explosion under water, oscillates with a period T proportional to $P^a*d^b*E^c$. Where `P' is the static pressure, `d' is the density of water, and `E' is the total energy of the explosion. Find the values of a, b, and c, and explain the physical reasoning behind your answer.\\
     \midrule
     \textbf{Example questions with long answer} \\
     \midrule
     (1) Analyze the impact of errors in data analysis on the validity of research findings, using the example of Reinhart and Rogoff's research on the relationship between debt and economic growth. How do such errors affect the development of economic policies, and what are the implications for the field of economics?\\
     (2) Prove that the relation $R = \{(1,2), (1,3), (1,4), (2,1), (2,3), (2,4), (3,1), (3,2), (3,4), (4,1), (4,1), $ $(4,3)\}$ is not transitive. Use the definition of transitivity to show that there exist elements $x_0, y_0, z_0$ such that $(x_0, y_0) \in R$ and $(y_0, z_0) \in R$ but $(x_0, z_0) \notin R$.\\
     (3) Astronomers use a new method to detect asteroids and measure their velocity. The method involves detecting energized electromagnetic waves at two different Earth times and using the relative motion of Earth with respect to the asteroid to calculate the velocity. Suppose the Earth spins about 360 degrees within 24 hours, and the asteroid moves in a straight path with respect to other stellar objects. If the angle between the flat Earth surface at point A0 and the direction of asteroid observable is ?, and the angle between the flat Earth surface at point A and the direction of asteroid observable is ?, derive an expression for the relative velocity of Earth with respect to the asteroid at position A0 and A. Use the relativistic Doppler formula to relate the frequencies of the electromagnetic waves detected at points A0 and A. Assume the velocity of the asteroid does not change within the time interval of two detections, and estimate the value of the asteroid's velocity. \\
     (4) Bob, a resident outside the US, has purchased a mobile app subscription from a California-based business for under \$200. However, due to a showstopper bug, the app is unusable for its main purpose. Bob has attempted to report the issue to the business's support team without success. Discuss the practicalities of Bob suing the California-based business from abroad, considering the requirements of Small Claims court in California and the potential application of consumer protection laws from Bob's home country. How might Bob's approach differ if he were to pursue litigation in a 'normal' court versus Small Claims court, and what are the implications of using an online store like Apple for the purchase?\\
     (5) Describe the differences in flow resistance between laminar and turbulent flows in a tube, and explain how the velocity profile changes in the transition from laminar to turbulent flow. Be sure to include the role of viscosity, density, and eddy viscosity in your answer.\\
\bottomrule
    \end{tabularx}
    \label{tab:example_questions}
\end{table*}

\section{Data Creation Details}

\subsection{Generation}
We use vllm for all generations. For annotating documents and  synthesizing questions, we use greedy decoding (i.e. temperature=$0$). For response generation for each question in \dataname, we use temperature=$0.7$ top\_p=$0.9$. Responses used in unsupervised self-training experiments are sampled using temperature=\{0, 0.5, 0.6, 0.7, 0.8, 0.9, 1.0, 1.2\} to encourage response diversity.

\subsection{SelfTrain-15k Curation}
\label{app:self_train_15k}
Specifically, for each question in the non-Diamond subset of GPQA, we retrieve the 1,000 most similar questions from \dataname, which were already decontaminated against the entire GPQA datset. Similarity is computed using cosine similarity between two question embeddings. After obtaining this candidate set, we apply deduplication and perform clustering, grouping the questions into 15,000 clusters. From each cluster, we select the questions closest to the cluster center, ensuring a diverse and representative dataset for downstream science reasoning tasks. This process resulted in a pool of 15,000 questions, which we refer to as \texttt{SelfTrain-15k}. 

% \newpage
\section{Human Evaluation of Question Quality}
\label{app:human_eval}
To assess question quality reliably, we also conducted a human evaluation by randomly sampling 100 questions from each dataset. Two expert annotators independently rated each question, and we used the average of their scores as the final quality measure. The results, shown in \autoref{tab:human_eval_q_quality}, confirm that the \dataname dataset consistently produces higher-quality questions.

\begin{table}[t]
\footnotesize
  \caption{Human evaluation of question quality across five datasets.}
  \label{tab:human_eval_q_quality}
\centering
\begin{tabular}{lccccc}
\toprule
Dataset               & MetaMathQA & OpenMathInstruct-2 & NuminaMath & WebInstruct & NaturalReasoning \\
\midrule
Quality score & 5.09       & 5.37               & 5.61       & 5.92        & 6.46            \\
\bottomrule
\end{tabular}
\end{table}

\section{Qwen2.5-7B Scaling Results}
\label{app:scale_qwen}
\begin{figure*}[t]
  \centering
  % left: figure ---------------------------------------------------------
  \begin{subfigure}[t]{0.49\textwidth}
    \vspace{0pt} % <─ ensures true top alignment
    \centering
    \includegraphics[width=\linewidth]{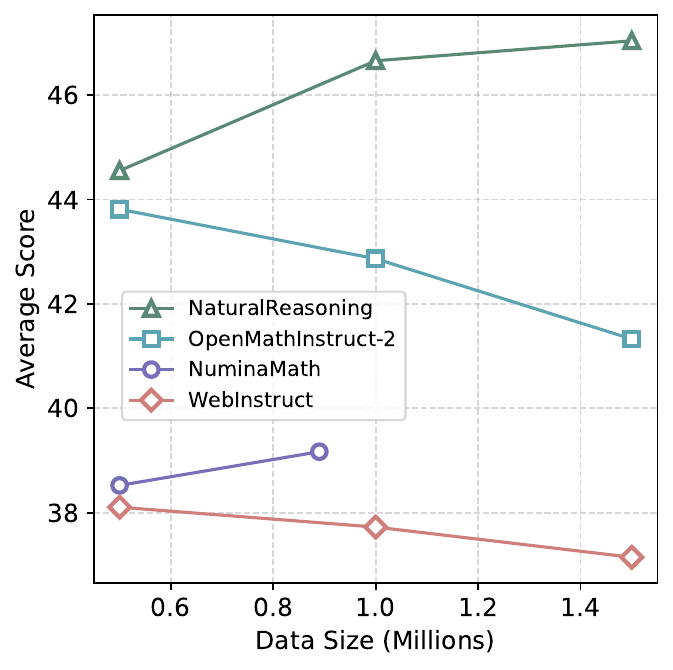}
    \caption{Average performance across MATH, GPQA, and MMLU-Pro when using varying sizes of data for training Qwen2.5-7B.
    }
    \label{fig:scaling_qwen}
  \end{subfigure}
  \hfill
  % right: table ---------------------------------------------------------
  \begin{subtable}[t]{0.49\textwidth}
    \vspace{0pt} % <─ same trick
\footnotesize
\setlength{\tabcolsep}{7pt}
\renewcommand{\arraystretch}{0.95}
\caption{Performance breakdown by benchmark, where the highest accuracy per data size is bolded.
% Math-specific datasets such as NuminaMath and OpenMathInstruct-2 perform well on MATH but poorly on on non-math specific tasks (GPQA and MMLU-Pro). \dataname generalizes well across all three tasks, suggesting high-quality and diverse training data.
}
\vspace{1mm}
\begin{tabular}{lccc}
\toprule
\textbf{SFT Dataset} & \textbf{500K} & \textbf{1M} & \textbf{1.5M} \\
\midrule
\multicolumn{4}{c}{\textit{MATH}} \\
\midrule
WebInstruct              & 38.82 & 40.12     &  38.49 \\
NuminaMath               & 62.68 & --     & --     \\
OpenMathInstruct-2               & 64.25 & 65.51  & 65.55  \\
NaturalReasoning     & \textbf{65.85} & \textbf{66.26} & \textbf{66.67} \\
\midrule
\multicolumn{4}{c}{\textit{GPQA}} \\
\midrule
WebInstruct              & \textbf{31.85} & \textbf{30.28}    &  \textbf{29.76} \\
NuminaMath               & 1.71 & --     & --     \\
OpenMathInstruct-2               & 18.23 & 14.21 & 9.08  \\
NaturalReasoning     & 16.44 & 20.76 &  21.35 \\
\midrule
\multicolumn{4}{c}{\textit{MMLU-Pro}} \\
\midrule
WebInstruct              & 43.65 & 42.77     &  43.18  \\
NuminaMath               & 51.18 & --     & --      \\
OpenMathInstruct-2               & 48.96 & 48.88 &  49.36  \\
NaturalReasoning     & \textbf{51.37} & \textbf{52.96} & \textbf{53.10} \\
\bottomrule
\end{tabular}
    
    \label{tab:qwen_scaling_breakdown}
  \end{subtable}

  \caption{Scaling results for Qwen2.5-7B model.}
  \label{fig:qwen_scaling_fig_tab}
\end{figure*}

The scaling trends for the Qwen2.5-7B model plotted by averaging performance on the three benchmarks are shown in \autoref{fig:scaling_qwen}. \autoref{tab:qwen_scaling_breakdown} provides a detailed breakdown of model performance across different dataset sizes and benchmarks. It is clear that \dataname shows superior scaling trends than other datasets.

\section{Reference Answer Usefulness}
\label{sec:ref_ans_use}
\subsection{Data Filtering In Knowledge Distillation}

% \wz{TODO:List example questions in the appendix for sample extraction for question with answer and without answer.}
We demonstrate the potential usefulness of reference answers using questions from \texttt{SelfTrain-15k}. We remove the questions that we are not able to extract a reference answer for and conduct a comparison to understand the utility of reference answers. We fine-tune the Llama3.1-8B-Instruct model using data filtered by final answer verification against a model trained on the unfiltered data.

% \subsection{Training Details}
For final answer verification, we use the prompt in Appendix \autoref{tab:prompt_self_score_with_ref} that prompts the model to judge whether the generated response using Llama3.3-70B-Instruct is in line with the reference final answer, using CoT reasoning. For training data filtering, we only keep the responses that have received a ``Yes'' final judgement. The training setup includes a learning rate of $1e^{-6}$, a batch size of 64, and training for three epochs, with checkpoints saved every 100 steps for the unfiltered experiment and 50 steps for the filtered experiment due to much smaller data size.

% \subsection{Results}
The results are shown in \autoref{tab:ref_ans_filter}. Filtering training data using reference answers leads to better performance despite a smaller training set. The filtered dataset contains 7,646 examples, significantly fewer than the 12,349 examples in the unfiltered dataset, yet achieves a higher score on both GPQA-Diamond (32.15 vs. 31.82) and MMLU-Pro (50.06 vs. 49.92). This suggests that higher-quality training data outweighs raw data quantity.

% The improvements are particularly notable given that the unfiltered data provides no measurable gain over the baseline on MMLU-Pro (31.82). This indicates that introducing lower-quality or noisy data can dilute training effectiveness, while filtering improves signal quality, leading to better model generalization. These results highlight the value of curating training data based on reference answers to enhance reasoning capabilities, even in non-mathematical domains.

\begin{table}[t]
\footnotesize
\setlength{\tabcolsep}{13pt}
\renewcommand{\arraystretch}{1.1}
\caption{
% SFT results using the reference answers to filter. We finetune Llama3.1-8B-Instruct on a subsample of unfiltered data from \dataname and a filtered version of this set, where questions are excluded if the distilled response from Llama3.3-70B-Instruct does not agree with the reference final answer. The filtered version performs better despite being much smaller, indicating the higher quality of our reference answers.
SFT results using reference answer filtering. We fine-tune Llama3.1-8B-Instruct on both an unfiltered subsample of \dataname and a filtered version, where questions are excluded if Llama3.3-70B-Instruct's response disagrees with the reference answer. Despite its smaller size, the filtered set performs better, highlighting the quality of our reference answers.
}
\label{tab:ref_ans_filter}
\begin{tabular}{lllll}
\toprule
           & \textbf{Training size}             & \textbf{GPQA-Diamond}                 & \textbf{MMLU-Pro} & \textbf{Average}                      \\

%            \midrule
% \multicolumn{4}{l}{\textit{Training Llama3.1-8B-Instruct}}     
% \\
\midrule
Llama3.1-8B-Instuct   & \multicolumn{1}{c}{--}                        & \multicolumn{1}{c}{31.82} & \multicolumn{1}{c}{49.79} & \multicolumn{1}{c}{40.81}  \\
Unfiltered SFT & \multicolumn{1}{c}{12,349} & \multicolumn{1}{c}{31.82} & \multicolumn{1}{c}{49.92} & \multicolumn{1}{c}{40.87}\\
Filtered SFT  & \multicolumn{1}{c}{7,646}  & \multicolumn{1}{c}{\textbf{32.15}} & \multicolumn{1}{c}{\textbf{50.06}} & \multicolumn{1}{c}{\textbf{41.11}} \\
\bottomrule
\end{tabular}
\end{table}

\subsection{Reinforcement Learning With Verifiable Rewards}

We conduct preliminary experiments applying reinforcement with verifiable rewards (RLVR) on the \dataname dataset. Specifically, we sample a subset of questions whose reference answers are shorter than 10 words and train Llama3.1-8B-Instuct using GRPO~\citep{shao2024deepseekmathpushinglimitsmathematical} with the General Verifier~\citep{general-reasoner} as the reward model. Training is performed with a batch size of 768. Despite only 50 optimization steps, the model already exhibit noticeable performance gains over the untrained baseline across multiple reasoning benchmarks as shown in \autoref{tab:ref_ans_grpo}.

\begin{table}[t]
\footnotesize
\setlength{\tabcolsep}{19pt}
\renewcommand{\arraystretch}{1.1}
\caption{
Online RL results using verifiable rewards on \dataname. We apply GRPO to Llama3.1-8B-Instruct using the General Verifier as the reward model.
}
\label{tab:ref_ans_grpo}
\begin{tabular}{lllll}
\toprule
   \textbf{Model}           & \textbf{MATH}             & \textbf{MMLU-Pro}                 & \textbf{GPQA} & \textbf{Average}                      \\

\midrule
Llama3.1-8B-Instuct   & \multicolumn{1}{c}{48.74}                        & \multicolumn{1}{c}{\textbf{49.79}} & \multicolumn{1}{c}{29.24} & \multicolumn{1}{c}{42.59}  \\
GRPO (Step 50) & \multicolumn{1}{c}{\textbf{52.20}} & \multicolumn{1}{c}{49.48} & \multicolumn{1}{c}{\textbf{31.47}} & \multicolumn{1}{c}{\textbf{44.39}}\\

\bottomrule
\end{tabular}
\end{table}

These early results suggest that reference answers in \dataname can be used in RLVR to further enhance reasoning performance, even with limited training. Due to current constraints in computational resources, we leave a more comprehensive exploration of RL-based fine-tuning methods for future work. 

\section{Evaluating NaturalReasoning Against OpenThoughts}
\label{app:compare_openthoughts}
To ensure a fair comparison, we adopt a similar knowledge distillation setup, using a stronger reasoning model (i.e., DeepSeek-R1) as the teacher and Llama-3.1-8B-Instruct as the student. Following the procedure in the OpenThoughts paper~\citep{guha2025openthoughtsdatarecipesreasoning}, we apply length filtering and then sample 100K questions from each dataset. As shown in \autoref{tab:compare_to_openthoughts}, the model trained on \dataname outperforms the one trained on OpenThoughts-114k on three out of four benchmarks.

\begin{table}[t]
\footnotesize
\setlength{\tabcolsep}{9pt}
\renewcommand{\arraystretch}{1.1}
\caption{
Knowledge distillation results comparing training on \dataname and OpenThoughts. We distill from DeepSeek-R1 into Llama3.1-8B-Instruct using 100K samples from each dataset after applying length filtering following~\citet{guha2025openthoughtsdatarecipesreasoning}. 
}
\label{tab:compare_to_openthoughts}
\begin{tabular}{lllll}
\toprule
   \textbf{Train Set}           & \textbf{GPQA-Diamond}             & \textbf{MATH-500}                 & \textbf{MMLU-Pro} & \textbf{SuperGPQA} \\

\midrule
OpenThoughts-114k (100K)   & \multicolumn{1}{c}{37.8}                        & \multicolumn{1}{c}{\textbf{82.2}} & \multicolumn{1}{c}{59.0} & \multicolumn{1}{c}{30.0}   \\
NaturalReasoning (100K)  & \multicolumn{1}{c}{\textbf{43.1}} & \multicolumn{1}{c}{69.0} & \multicolumn{1}{c}{\textbf{61.2}} & \multicolumn{1}{c}{\textbf{32.2}} \\

\bottomrule
\end{tabular}
\end{table}

These results show that \dataname achieves stronger performance on general reasoning and science benchmarks such as GPQA-Diamond, MMLU-Pro, and SuperGPQA~\citep{pteam2025supergpqascalingllmevaluation}, indicating that it complements existing reasoning datasets like OpenThoughts by providing broader coverage and better generalization.

\section{Cross-Domain Generalization of NaturalReasoning}
\label{app:cross_domain_generalization}

While our main experiments focus on reasoning benchmarks, we also conduct preliminary studies to examine \dataname’s applicability to broader knowledge domains. Specifically, we train Llama-3.1-8B-Instruct on 100K randomly sampled \dataname examples, using DeepSeek-R1 as the teacher model for knowledge distillation. The resulting model is then evaluated on SuperGPQA~\citep{pteam2025supergpqascalingllmevaluation}, which includes diverse subjects beyond mathematics and science—such as philosophy, law, history, economics, literature, and sociology. The results on non-reasoning categories are shown in \autoref{tab:generalization_to_non_reasoning}.

\begin{table}[t]
\footnotesize
\setlength{\tabcolsep}{3pt}
\renewcommand{\arraystretch}{1.1}
\caption{
Cross-domain evaluation results after training on \dataname. We fine-tune Llama3.1-8B-Instruct on 100K \dataname examples with DeepSeek-R1 as the teacher model and evaluate on SuperGPQA’s non-reasoning subjects.
}
\label{tab:generalization_to_non_reasoning}
\begin{tabular}{lllllllll}
\toprule
   \textbf{Model}     & \textbf{Philosophy}      & \textbf{Economics}      & \textbf{History} & \textbf{Education} & \textbf{Law}  & \textbf{Management}  & \textbf{Literature} & \textbf{Sociology} \\

\midrule
Baseline  & \multicolumn{1}{c}{29.36} & \multicolumn{1}{c}{23.17} & \multicolumn{1}{c}{17.24} & \multicolumn{1}{c}{26.03} & \multicolumn{1}{c}{\textbf{29.63}} & \multicolumn{1}{c}{29.54} & \multicolumn{1}{c}{23.37} & \multicolumn{1}{c}{23.78} \\
Train on NR & \multicolumn{1}{c}{\textbf{32.56}} & \multicolumn{1}{c}{\textbf{34.86}} & \multicolumn{1}{c}{\textbf{27.49}} & \multicolumn{1}{c}{\textbf{30.17}} & \multicolumn{1}{c}{\textbf{29.63}} & \multicolumn{1}{c}{\textbf{31.14}} & \multicolumn{1}{c}{\textbf{27.08}} & \multicolumn{1}{c}{\textbf{33.57}}\\

\bottomrule
\end{tabular}
\end{table}

Despite its focus on reasoning-centric tasks, \dataname improves performance across a wide range of non-reasoning domains. These results highlight its potential as a versatile foundation for training models with broad domain generalization.

\section{Prompts}
\label{app:prompts}
% The prompt we used for annotating reasoning from the document, generating a question, and reference answer are shown in \autoref{tab:prompt_annotate_reasoning_part1} and \autoref{tab:prompt_annotate_reasoning_part2}.

The prompt we used for annotating reasoning from the document is shown in \autoref{tab:prompt_annotate_reasoning_part1}, \autoref{tab:prompt_annotate_reasoning_part2}.
We additionally provide the prompt for annotating question validity and difficulty (\autoref{tab:prompt_annotate_quality}), the prompt used to check if a generated response matches the reference (\autoref{tab:prompt_self_score_with_ref}), and the prompt for self scoring (\autoref{tab:prompt_self_score}).

\begin{figure}[th!]
\centering
\small
\begin{tcolorbox}[colback=green!10!white, % Background color
                  colframe=green!30!white, % Frame color
                  width=1\textwidth, % Width of the tcolorbox
                  arc=4mm, % Radius of the rounded corners
                  auto outer arc,
                   sharp corners=south, % Ensure corners are well-rounded
                  fontupper=\ttfamily, % Use monospaced font for readability
                  before upper={\setlength{\parindent}{0em}} % Ensure consistent indentation
                  ]
\begin{lstlisting}[basicstyle=\ttfamily\tiny, breaklines=true,mathescape=true]
Evaluate the text below according to the scoring instruction and criteria. If the scores are high on each axis, derive an exam question following the instructions.

## Scoring Instruction
1. Evaluate the text on each criteria step by step. Provide your honest answer to each sub-question. If the answer to a sub-question is a confident Yes, add or subtract the points corresponding to the criteria.
2. Keep track of the running points from each criteria to get the total score.
3. Summarize your final evaluation results in a valid JSON object following the instruction below.

### Scoring Criteria

**Criteria 1: Problem Completeness**
* The content does not have clear main question, or enough clues to derive the correct answer. (0 point)
* The content includes a main question, and enough clues to derive the correct answer. (+1 point)
* The text shows evidence of engagement and discussion among multiple authors, including proposing answers, evaluating and reflecting on answers, responding to critiques, revising and editing answers. (+1 point)

**Criteria 2: Problem Complexity and Technical Depth**

* The difficulty of the content is college-level or below. (0 point)
* The difficulty of the content is graduate-level or above, and only domain experts can understand. (+1 point)
* The question being discussed is so challenging that even highly skilled non-experts would not be able to fully understand the question or provide a correct answer, even after spending 30 minutes searching the internet or reading up literature. (+1 point)

**Criteria 3: Technical Correctness and Accuracy**

* The text contains significant technical errors or inaccuracies. (-1 point)
* The text demonstrates some technical correctness, but with notable flaws or omissions (e.g., incorrect units, incomplete derivations). (0 point)
* The text demonstrates technical correctness, with some minor flaws or omissions (e.g., minor algebraic errors, incomplete explanations). (+0.5 point)
* The text demonstrates high technical correctness, with clear and accurate explanations (e.g., precise definitions, complete derivations). (+0.5 point)
* The text exemplifies exceptional technical correctness, with rigorous and precise explanations (e.g., formal proofs, precise calculations). (+1 point)

**Criteria 4: Thinking and Reasoning**

* The text lacks any evidence of thinking or reasoning. (-1 point)
* The text demonstrates some basic thinking and reasoning (+0.5 point), such as:
        + A straightforward application of a known technique.
        + A simple analysis of a problem.
* The text demonstrates some thinking and reasoning (+0.5 point), such as:
        + A consideration of multiple approaches to a problem.
        + A discussion of the trade-offs between different solutions.
* The text demonstrates significant thinking and reasoning (+1 points), such as:
        + Multi-step reasoning chains to solve a complex problem.
        + Advanced reasoning patterns often used in specialized science domains.
* The text exemplifies exceptional thinking and reasoning (+1 points), such as:
        + A highly innovative and creative approach to solving a complex problem in specialized domains.
        + Combining multiple reasoning and thinking techniques, with novel abstraction of the problem.

## Instruction on Exam Question and Final Report
* Step 1. If BOTH Criteria 1 and Criteria 2 scores above zero, transform the original question being discussed to an exam question. The question should focus on problem-solving and there should exist a correct answer to the question. The question should be descriptive, i.e. use the details and notations from the original text as much as possible. The question must be self-contained, concrete, well-defined, i.e. it should NOT contain any missing information nor should it contain any ambiguity or subjectiveness.
* Step 2. If BOTH Criteria 1 and Criteria 3 scores above zero, determine whether the text contains a correct solution to the question or not. If the discussion DOES contain a correct solution, try to extract the gists and important details from the correct answer. Then use those key information to derive a correct answer to the question. If there is a single final answer, conclude the correct answer with: "Therefore, the final answer is: \\boxed{{answer}}.", where [answer] is the number or expression that is the final answer.
* Step 3. If BOTH Criteria 2 and Criteria 4 have non-zero scores, write down a list of critical knowledge and reasoning steps which are required to derive a correct answer to the exam question. Each item in the list must be descriptive, specific and concrete.
* Step 4. Label the question difficulty with Easy, Medium, Hard, and Extra Hard.


\end{lstlisting}

\end{tcolorbox}
\caption{Prompt for annotating reasoning from the document, generating a question and reference answer. (Part 1)}
\label{tab:prompt_annotate_reasoning_part1}
\end{figure}

\begin{figure}[th!]
\centering
\small
\begin{tcolorbox}[colback=green!10!white, % Background color
                  colframe=green!30!white, % Frame color
                  width=1\textwidth, % Width of the tcolorbox
                  arc=4mm, % Radius of the rounded corners
                  auto outer arc,
                   sharp corners=south, % Ensure corners are well-rounded
                  fontupper=\ttfamily, % Use monospaced font for readability
                  before upper={\setlength{\parindent}{0em}} % Ensure consistent indentation
                  ]
\begin{lstlisting}[basicstyle=\ttfamily\tiny, breaklines=true,mathescape=true]

Finally, copy all your analysis from above into a JSON object at the end of the final report. The JSON object should contain the following attributes:
- `"scores"`: a list of dictionary entries, where each entry contains the criteria name and corresponding score on that criteria.
- `"exam\_question"`: a string recording the full exam question derived from Step 1 analysis. DO NOT omit any details. If the scores for Criteria 1 and Criteria 2 are low and no exam question can be made out of the text, return an empty string.
- `"correct\_answer"`: a string recording the correct answer derived from Step 2. If the discussion does not contain a correct answer, return an empty string.
- `"knowledge\_and\_reasoning\_steps"`: a list of strings, where each entry copying the critical piece of knowledge or important reasoning steps derived from Step 3. If either Criteria 2 or Criteria 4 has score zero, return an empty list.
- `"question\_difficulty"`: a string recording the difficulty of the question derived from Step 4.

### Text
{text}

\end{lstlisting}

\end{tcolorbox}
\caption{Prompt for annotating reasoning from the document, generating a question and reference answer. (Part 2)}
\label{tab:prompt_annotate_reasoning_part2}
\end{figure}

\begin{figure}[th!]
\centering
\small
\begin{tcolorbox}[colback=green!10!white, % Background color
                  colframe=green!30!white, % Frame color
                  width=1\textwidth, % Width of the tcolorbox
                  arc=4mm, % Radius of the rounded corners
                  auto outer arc,
                  sharp corners=south, % Ensure corners are well-rounded
                  fontupper=\ttfamily, % Use monospaced font for readability
                  before upper={\setlength{\parindent}{0em}} % Ensure consistent indentation
                  ]

\begin{lstlisting}[basicstyle=\ttfamily, breaklines=true]
Your task is to verify and improve the quality of a question. 

A valid question must meet the following criteria:
* The question should contain a problem to be solved, instead of only presenting statements.
* The question should be well-defined and self-contained, i.e. have all the necessary information to derive an answer.
* The question should be specific and clear. There should be one correct answer to the question. 
* The question should not refer to external resources, such as figures, videos, etc.
* To derive an answer, multi-step reasoning and recalling relevant knowledge is required.
* The difficulty should be graduate-level or above.
* The question can contain LaTex but it should be correct.
If a question does not meet any of the criterion above, revise it till it meets all the criteria.

IMPORTANT: Put your final answer in a JSON object with two fields:
- "question_quality_score": rate how well the question meets all the criteria, on a scale of 1 to 10.
- "improved_question": revised question which will meet the criteria better and thus has a higher question quality score. 

Question:
{question}
\end{lstlisting}
\end{tcolorbox}
\caption{Prompt for annotating quality scores.}
\label{tab:prompt_annotate_quality}
\end{figure}

% \clearpage

\begin{figure}[th!]
\centering
\small
\begin{tcolorbox}[colback=green!10!white, % Background color
                  colframe=green!30!white, % Frame color
                  width=0.99\textwidth, % Width of the tcolorbox
                  arc=4mm, % Radius of the rounded corners
                  auto outer arc,
                  sharp corners=south, % Ensure corners are well-rounded
                  fontupper=\ttfamily, % Use monospaced font for readability
                  before upper={\setlength{\parindent}{0em}} % Ensure consistent indentation
                  ]

\begin{lstlisting}[basicstyle=\ttfamily, breaklines=true]
You are an expert evaluator tasked with deciding whether a response meets the standards of quality and correctness for general reasoning tasks. Your evaluation must consider both the quality of the reasoning process (chain of thought, CoT) and the correctness or appropriateness of the final answer.

### Evaluation Criteria:
1. **Correctness of the Final Answer**: 
   - Does the final answer align with the reference answer or the expected outcome?
2. **Quality of the Thinking Process (CoT)**:
   - Is the reasoning logical, coherent, and free from significant errors?
   - Does the reasoning support the final answer in a clear and step-by-step manner?
3. **Completeness**: 
   - Does the response adequately address all aspects of the instruction or problem?

### Input Details:
- **Instruction**: {Describe the task, problem, or instruction here.}
- **Reference Answer**: {Provide the expected or ideal outcome here.}
- **Response**: {Include the response to be evaluated, containing both the CoT and the final answer.}

### Task:
Analyze the response provided and decide if it is a "Yes" or "No" based on the following:
- **"Yes"**: The response meets the required standards for correctness, reasoning, and completeness.
- **"No"**: The response fails to meet one or more of the standards.

Provide a brief explanation of your decision, highlighting specific strengths or weaknesses in the reasoning process (CoT), the final answer, or completeness.

**Response Format**:
1. **Explanation**:  
   - **Final Answer Evaluation**: (Discuss correctness and consistency.)  
   - **Chain of Thought Evaluation**: (Discuss logic and coherence.)  
   - **Completeness**: (Assess whether the response fully addresses the instruction.)  
2. **Judgment**: Yes/No  
---

[FEW-SHOT EXAMPLES HERE]
---

### Current Input:
**Instruction**: <INSTRUCTION>
**Reference Answer**: <ANSWER>
**Response**:  
<RESPONSE>

**Evaluation**:

\end{lstlisting}
\end{tcolorbox}
\caption{Prompt used to check if a response matches the reference answer.}
\label{tab:prompt_self_score_with_ref}
\end{figure}

\begin{figure}[th!]
\centering
\small
\begin{tcolorbox}[colback=green!10!white, % Background color
                  colframe=green!30!white, % Frame color
                  width=1\textwidth, % Width of the tcolorbox
                  arc=4mm, % Radius of the rounded corners
                  auto outer arc,
                  sharp corners=south, % Ensure corners are well-rounded
                  fontupper=\ttfamily, % Use monospaced font for readability
                  before upper={\setlength{\parindent}{0em}} % Ensure consistent indentation
                  ]

\begin{lstlisting}[basicstyle=\ttfamily, breaklines=true]
You are an expert evaluator tasked with analyzing a response to a general reasoning problem. Your goal is to determine if the response demonstrates good reasoning (CoT) and whether the reasoning makes sense overall.

### **Evaluation Criteria**:
1. **Reasoning Quality (CoT)**:
   - Does the reasoning follow a logical and coherent sequence?
   - Are the steps valid and free of major errors?
   - Does the reasoning align with standard problem-solving practices?

2. **Accuracy**:
   - Does the reasoning lead to the correct or valid conclusion based on the instruction?

3. **Clarity**:
   - Is the response clear and easy to understand?

### Input Details:
- **Instruction**: {Describe the task, problem, or instruction here.}
- **Response**: {Include the response to be evaluated, containing the chain of thought (CoT).}

### **Task**:
Analyze the response and decide if it meets the standards for correctness, reasoning quality, and clarity. Provide your judgment as either **"Yes"** (the response is good) or **"No"** (the response is not good). Then, briefly explain your decision.

**Response Format**:
1. **Judgment**: Yes/No  
2. **Explanation**:  
   - **Reasoning Quality (CoT)**: (Assess the reasoning process in detail.)  
   - **Accuracy**: (Evaluate whether the reasoning leads to the correct conclusion.)  
   - **Clarity**: (Comment on the clarity of the response.)  

---
[FEW-SHOT EXAMPLES]

---

### Current Input:
**Instruction**: <INSTRUCTION>
**Response**:  
<RESPONSE>

**Evaluation**:
1. **Judgment**: 

\end{lstlisting}
\end{tcolorbox}
\caption{Prompt for self scoring.}
\label{tab:prompt_self_score}
\end{figure}

\end{document}